\documentclass{article}

\PassOptionsToPackage{numbers,square,sort}{natbib}

\usepackage[utf8]{inputenc} 
\usepackage[T1]{fontenc}    
\usepackage{hyperref}       
\usepackage{url}            
\usepackage{booktabs}       
\usepackage{amsfonts}       
\usepackage{nicefrac}       
\usepackage{microtype}      
\usepackage{xcolor}         
\usepackage{graphicx}
\usepackage{subcaption}
\usepackage{caption}
\usepackage{wrapfig}
\usepackage{hyphenat}
\usepackage{setspace}
\usepackage{amsmath}
\usepackage{amssymb}
\usepackage{bbm}
\usepackage{mathtools}
\usepackage{amsthm}
\usepackage{longtable}
\usepackage{multirow}
\usepackage{array}
\usepackage{colortbl}
\usepackage{float}
\usepackage{tabularx}
\usepackage{enumitem}
\usepackage{threeparttable}
\usepackage{titletoc}

\usepackage[preprint]{my_style}
\setcitestyle{numbers,square,comma}

\usepackage{tcolorbox}
\tcbuselibrary{skins, breakable, listings}
\usepackage{simpleicons}

\definecolor{brandblue}{RGB}{57,95,207}
\definecolor{linkblue}{HTML}{0064E0}
\definecolor{textgray}{HTML}{1C2B33}
\definecolor{boxbg}{HTML}{F1F4F7}
\definecolor{gaingreen}{RGB}{0,128,0}
\definecolor{lossred}{RGB}{200,0,0}
\definecolor{oursblue}{RGB}{222,235,247}
\definecolor{githubblack}{HTML}{181717}
\definecolor{huggingyellow}{HTML}{FFD21E}

\newcommand{\imp}[1]{\textcolor{gaingreen}{#1}}
\newcommand{\reg}[1]{\textcolor{lossred}{#1}}

\newcommand{\method}{IterCAD}

\hypersetup{
  colorlinks=true,
  linkcolor=brandblue,
  citecolor=brandblue,
  urlcolor=linkblue
}

\newcommand{\paperTitle}{IterCAD: An Iterative Multimodal Agent for Visually-Grounded CAD Generation and Editing}

\newcommand{\paperAuthors}{%
  {\sffamily\bfseries Tao Hu$^{1, 2, *}$}, 
  {\sffamily\bfseries Jiaxin Ai$^{1, 3, 4, *}$}, 
  {\sffamily\bfseries Licheng Wen$^{1, 3, *}$}, 
  {\sffamily\bfseries Xueheng Li$^{2, *}$}, 
  {\sffamily\bfseries Shu Zou$^{1, 5}$}, 
  {\sffamily\bfseries Siqi Li$^{1, 6}$}, \\
  {\sffamily\bfseries Nianchen Deng$^{1}$}, 
  {\sffamily\bfseries Xinyu Cai$^{1}$}, 
  {\sffamily\bfseries Hongbin Zhou$^{1}$}, 
  {\sffamily\bfseries Pinlong Cai$^{1}$}, 
  {\sffamily\bfseries Daocheng Fu$^{1, 7}$}, \\
  {\sffamily\bfseries Yu Yang$^{1, 6}$}, 
  {\sffamily\bfseries Hairong Zhang$^{1, 7}$}, 
  {\sffamily\bfseries Botian Shi$^{1, 3}$}, 
  {\sffamily\bfseries Xuemeng Yang$^{1, \dagger}$}
}

\newcommand{\paperAffiliations}{%
  {\normalsize $^1$ Shanghai Artificial Intelligence Laboratory} \
  {\normalsize $^2$ University of Science and Technology of China} \\
  {\normalsize $^3$ Shanghai Innovation Institute} \
  {\normalsize $^4$ Wuhan University} \
  {\normalsize $^5$ The Australian National University} \\
  {\normalsize $^6$ Zhejiang University} \
  {\normalsize $^7$ Fudan University}
}

\newcommand{\paperNotes}{%
  {\small $^*$ Equal contribution. \quad $^\dagger$ Corresponding author}%
}

\newcommand{\githubLink}{https://github.com/KnowledgeXLab/IterCAD}
\newcommand{\hfModelLink}{https://huggingface.co/KnowledgeXLab/IterCAD-RL}
\newcommand{\hfDataLink}{https://huggingface.co/datasets/KnowledgeXLab/IterCAD_Data}
\newcommand{\publishDate}{\today}
\newcommand{\coverSimpleIcon}[2][black]{\raisebox{-0.12em}{\textcolor{#1}{\fontsize{9}{11}\selectfont\simpleicon{#2}}}}

\newcommand{\renderFrontBox}{%
    \tcbset{
    enhanced, frame hidden,
    colback=boxbg,
    left=0.5cm, right=0.5cm, top=0.5cm, bottom=0.5cm,
    arc=16pt,
    before skip=0pt,
    grow to left by=1.5pt, grow to right by=1.5pt,
    overlay={

    \node[anchor=north east, at=(frame.north east), xshift=-2.3cm, yshift=-0.5cm] 
        {\includegraphics[height=1cm]{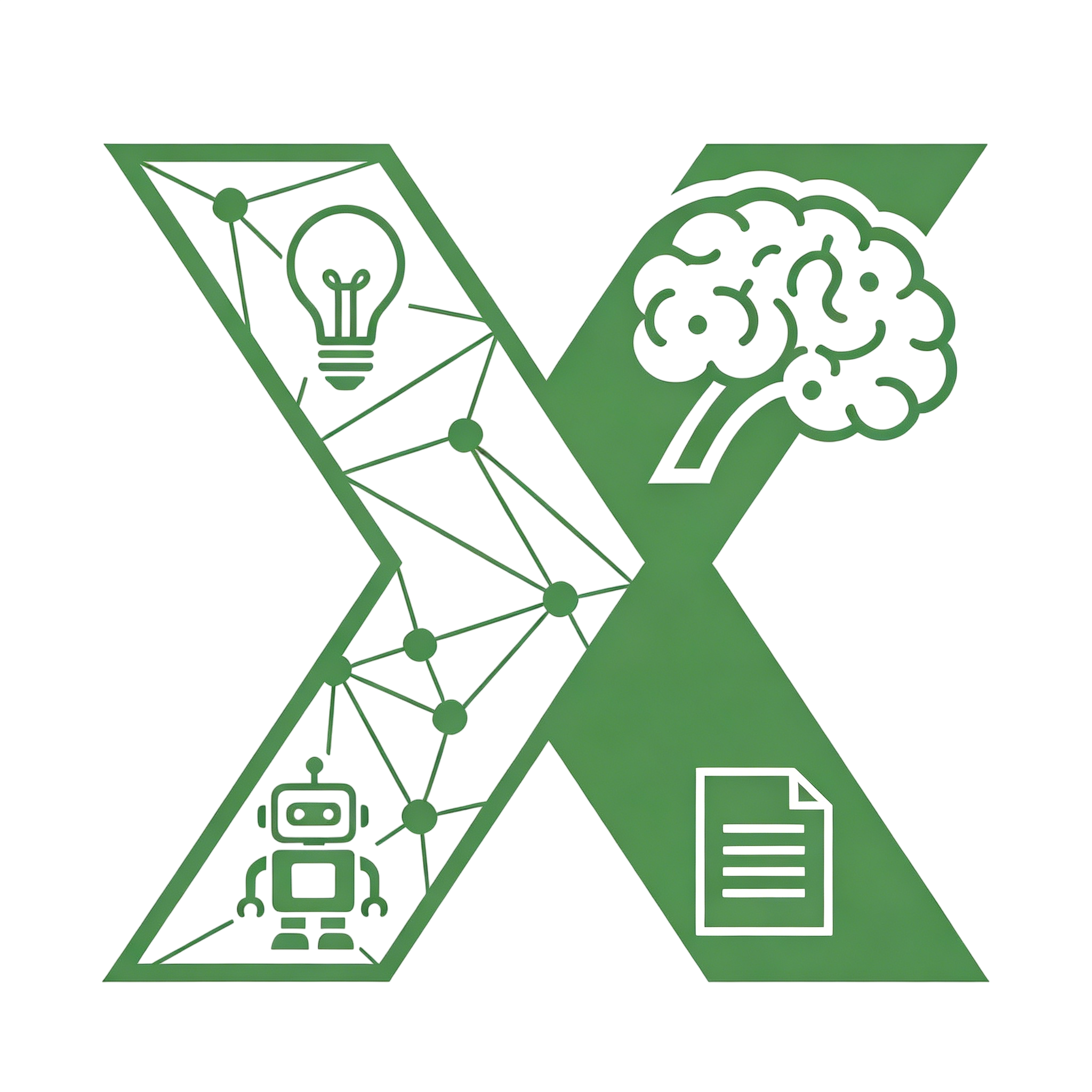}};
    \node[anchor=north east, at=(frame.north east), xshift=-0.5cm, yshift=-0.5cm] 
        {\includegraphics[height=1cm]{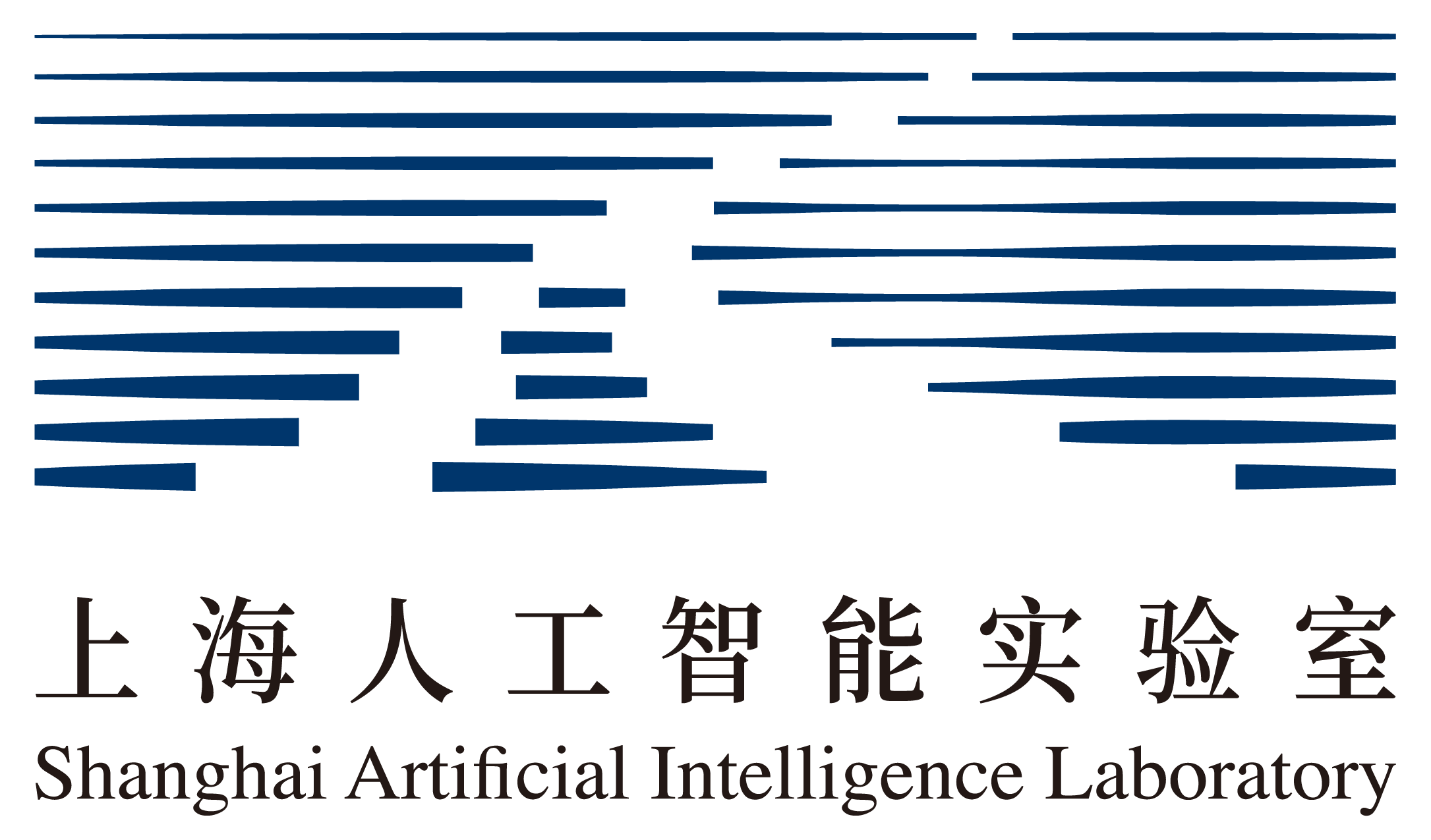}};
    }
  }%
  \begin{tcolorbox}
    \setlength{\parindent}{0cm}
    \setlength{\parskip}{0.5cm}
    {
      \setlength{\parskip}{0cm}
      \raggedright
      \nohyphens
      {
        \vskip 1.25cm
        {\fontsize{18}{24}\selectfont\sffamily\bfseries\textcolor{black}{%
          \raisebox{-0.15em}[0pt][0pt]{\includegraphics[height=2.05em]{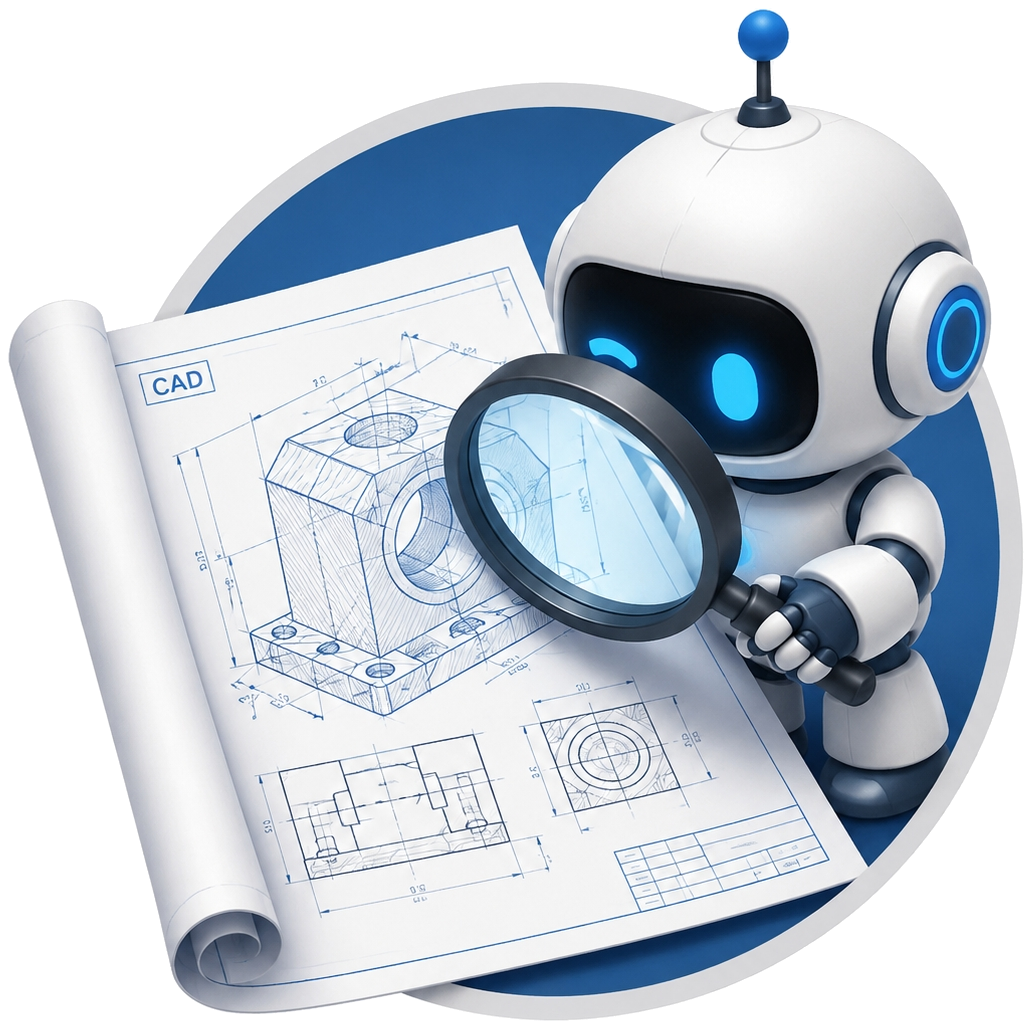}}%
          \kern0.04em
          \paperTitle}\par}
      }
      \vskip 0.35cm
      \paperAuthors\par
      \vskip 0.35cm
      \paperAffiliations\par
      \vskip 0.2cm
      \paperNotes\par
    }
    {\color{textgray}%
    Computer-Aided Design is pivotal in modern manufacturing, yet existing automated methods predominantly rely on open-loop, one-shot generation, creating a mismatch with iterative real-world practices.
In this paper, we present \method{}, a unified multimodal agent framework for closed-loop, interactive CAD generation and editing.
We formulate the task as a multi-turn interaction between a multimodal agent and an executable CAD sandbox, covering three tasks: Drawing-to-Code, Text-to-Code, and Interactive Editing.
To support this, we develop a data synthesis pipeline incorporating advanced industrial manufacturing features to generate standard-compliant multi-view engineering drawings, complex code-editing tasks, and high-fidelity interaction trajectories.
We optimize the agent via progressive SFT followed by geometry-aware reinforcement learning with viable-prefix masking to enhance code executability and geometric fidelity.
Finally, we introduce the IterCAD-Bench evaluation suite and propose the Chamfer Distance Tolerance-Recall (CD-TR) curve alongside its AUC-TR metric, establishing a survivor-bias-free standard that unifies code validity and geometric precision.
Extensive experiments demonstrate that \method{} achieves highly competitive performance across multiple benchmarks, significantly outperforming existing approaches in both code executability and geometric precision, while exhibiting superior capabilities in closed-loop iterative refinement.

\par}
    \vskip 0.4cm
    {
      \setlength{\parskip}{0cm}
      {\small {\sffamily\bfseries \raisebox{-0.2em}{\includegraphics[width=0.025\linewidth]{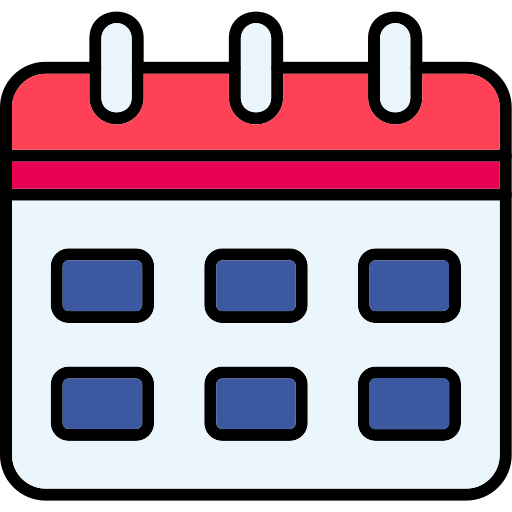}}~~Date:} \publishDate}\\[0.28em]
      {\small\sffamily\bfseries
        \href{\githubLink}{\coverSimpleIcon[githubblack]{github}~~Code}\quad
        \href{\hfModelLink}{\coverSimpleIcon[huggingyellow]{huggingface}~~Model}\quad
        \href{\hfDataLink}{\coverSimpleIcon[huggingyellow]{huggingface}~~Data}}\par
    }
  \end{tcolorbox}
  \tcbset{reset}
}

\begin{document}

\newgeometry{top=1in, bottom=0.75in, textwidth=6.3in, textheight=9in}
\renderFrontBox

\section{Introduction}

Computer-Aided Design (CAD) plays a central role in engineering and manufacturing~\cite{briere2012comparing}, yet constructing high-quality parametric models remains labor-intensive. Recent advances in Multimodal Large Language Models (MLLMs), together with programmatic CAD frameworks such as CadQuery~\cite{cadquery240} and build123d\footnote{\url{https://github.com/gumyr/build123d}}, have enabled a new paradigm for generating executable parametric CAD programs from text, images, and engineering drawings~\cite{xie2025text,li2024cad,yuan2026clarify,xu2022skexgen,barkley2026cadsmith,niu2025cme}. While recent programmatic methods (e.g., CAD-Coder~\cite{guan2026cad} and CADrille~\cite{kolodiazhnyi2025cadrille}) demonstrate the strong potential in language-model-driven CAD automation by generating executable Python code to procedurally construct 3D models, they are prone to minor geometric inconsistencies or topological defects during synthesis, often leading to runtime errors or visual artifacts~\cite{kabisov2026cadreasoner}.

\begin{figure}[t]
  \centering
  \includegraphics[width=0.66\columnwidth]{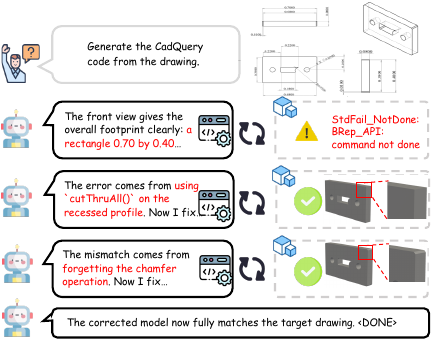}
  \caption{\method{} mimics the human ``generate--verify--refine'' workflow. Guided by multi-view engineering drawings, the agent iteratively optimizes CadQuery code by debugging execution failures and rectifying geometric defects, achieving seamless alignment with the target design.}
  \label{fig:Motivation}
  \vspace{-15pt}
\end{figure}

The fundamental limitation of current CAD generation stems from a critical mismatch between professional, iterative ``generate–verify–refine'' human workflows and prevailing ``open-loop, one-shot'' paradigms that directly synthesize complete programs without intermediate self-correction. Although recent multi-turn frameworks~\cite{zhou2026cad,kabisov2026cadreasoner,giannone2026gift} attempt to bridge this gap, their effectiveness is severely bottlenecked by coarse and shallow feedback signals: point-cloud metrics only capture global geometric discrepancies without localizing faulty parameters, while compiler feedback merely verifies syntactic executability while ignoring underlying topological logic, causing these systems to degenerate into unguided trial-and-error. Furthermore, existing benchmarks~\cite{khan2024text2cad,wu2021deepcad} misalign with real-world practices by overrepresenting simplistic "sketch-and-extrude" patterns~\cite{li2023secad} over advanced features (e.g., fillets, shells, chamfers) and prioritizing static generation over iterative editing. Crucially, current evaluation protocols suffer from a severe "survivor bias" by calculating geometric metrics exclusively on successfully executed codes, thereby failing to faithfully quantify an agent's true error recovery and robustness under realistic engineering settings.

To address these workflow and evaluation gaps, we introduce \method{}, a drawing-driven CAD agent governed by a ``Look and Loop'' philosophy shown in Figure~\ref{fig:Motivation}. Moving beyond static, one-shot synthesis, \method{} formulates CAD modeling as a closed-loop iterative process. Specifically, \textbf{Look} leverages multi-view engineering drawings with dimensional constraints as persistent references to isolate defects and maintain geometric consistency. Concurrently, \textbf{Loop} establishes a multi-dimensional refinement mechanism integrating compiler, execution, and visual feedback, empowering the agent to incrementally optimize designs through targeted ``generate--verify--refine'' cycles rather than blind regeneration.

To enable these closed-loop capabilities, we introduce a two-stage training recipe combined with a robust evaluation framework tailored for interactive CAD modeling.
We first perform a progressive cold-start supervised fine-tuning (SFT) on high-quality multi-turn interaction trajectories to initialize procedural CAD reasoning across generation, editing and refinement under complex engineering operations.
This is followed by reinforcement learning (RL) integrating geometry-aware rewards and Geometry-Viable Prefix Masking (GVPM) to encourage robust iterative correction and executable consistency, transforming static CAD generation into an interactive, self-correcting design process that continually refines programs based on execution outcomes, visual observations, and modification intents.
We also introduce IterCAD-Bench, a comprehensive evaluation suite spanning both interactive CAD generation (\textit{IterCAD-Draw}) and local editing scenarios (\textit{IterCAD-Edit}).
To eliminate the severe survivor bias of existing protocols, we propose the Chamfer Distance Tolerance-Recall (CD-TR) Curve and its scalar counterpart, Area Under the CD-TR Curve (AUC-TR); by explicitly accounting for failed generations ($CD\to\infty$), this joint metric faithfully measures both executability and geometric fidelity across the entire test distribution without performance distortions.

Our contributions are as follows:

\begin{itemize}[leftmargin=*, itemindent=0pt, itemsep=0pt, topsep=2pt]
  \item We introduce the \method{}, a drawing-driven agent framework that unifies CAD generation and editing into a closed-loop, multi-turn synthesis process by leveraging multi-view engineering drawings as spatial anchors.
  \item We design a two-stage training recipe that combines a progressive cold-start SFT stage with a geometry-aware RL stage enhanced by geometry-viable prefix masking, explicitly instilling robust self-correction and achieving competitive performance on multiple industrial CAD benchmarks.
  \item We construct IterCAD-Bench, a comprehensive multimodal dataset covering diverse engineering operations, and introduce the CD-TR curve to establish a survivor-bias-free evaluation standard that jointly quantifies execution validity and geometric precision.
\end{itemize}

\section{Related Work}

Learning-based 3D CAD generation fundamentally relies on data representations, which can be categorized into five distinct paradigms: (1) \emph{CSG-based methods}~\cite{du2018inversecsg, yu2022capri} using Boolean operations; (2) \emph{B-rep methods}~\cite{jayaraman2022solidgen, xu2024brepgen} predicting boundary networks; (3) \emph{Sequence-based methods}~\cite{wu2021deepcad, khan2024text2cad,govindarajan2025cadmium} encoding tokenized command lines; (4) \emph{Structured-text paradigms}~\cite{li2025cad, li2025seek} utilizing customized lightweight syntax; and (5) \emph{Code-based frameworks}~\cite{niu2026intent, guan2026cad, kolodiazhnyi2025cadrille} generating fully parametric scripts (e.g., CadQuery) to maximize editing flexibility and LLM compatibility. Despite these programmatic representations, conventional frameworks remain fundamentally constrained by an open-loop, one-shot paradigm that lacks iterative self-correction, while existing synthetic benchmarks~\cite{willis2021fusion, khan2024text2cad} remain heavily dominated by trivial sketch-and-extrude loops. To transition toward interactive refinement, recent multi-turn agentic systems~\cite{yuan2025cad, an2026pr, fan2025caddesigner, gong2026toolcad} couple LLMs with execution environments to iteratively repair design defects. However, their supervisory feedback remains critically coarse and ambiguous: environment-based compiler feedback~\cite{zhou2026cad} verifies syntactic validity but is blind to underlying geometric logic, while distance-based spatial feedback~\cite{kabisov2026cadreasoner} (e.g., Chamfer Distance on point clouds) only captures global shape errors. Crucially, these spatial deviations fail to localize errors or attribute them to specific 2D sketch dimensions or topological entities, frequently causing multi-turn iterations to degenerate into an unguided blind search. A more comprehensive analysis of these CAD paradigms and multi-turn feedback mechanisms is detailed in Appendix~\ref{app:related_work}.

\section{Methodology}
\label{sec:methodology}

In this section, we present \textbf{\method{}}, a unified agent framework for interactive CAD generation and editing. We first formulate the task as a multi-turn agent-environment interaction in Sec.~\ref{subsec:problem_formulation}. We then describe the CAD data curation pipeline in Sec.~\ref{subsec:dataset}. Sec.~\ref{subsec:training} introduces our two-stage training strategy, consisting of cold-start SFT and geometry-aware RL. Finally, Sec.~\ref{subsec:benchmark} exhibits the constructed benchmark for evaluating multi-turn CAD generation and editing, along with the \textit{CD-TR Curve}, a novel metric designed to eliminate survivor bias by unifying code validity and geometric fidelity.

\begin{figure*}[t]
  \centering
  \includegraphics[width=0.98\linewidth]{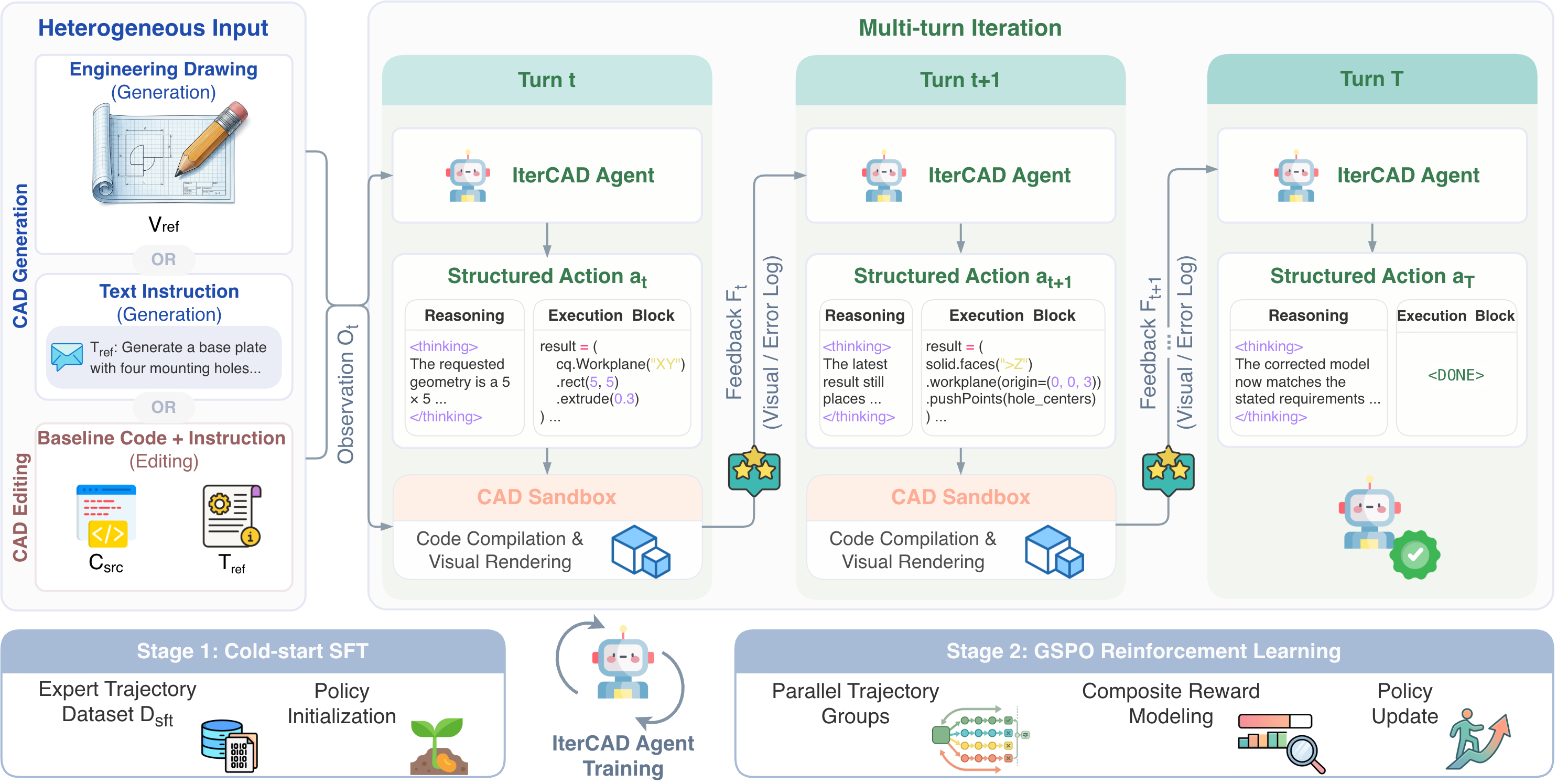}
  \caption{Overview of the \method{} framework. \method{} formulates interactive CAD generation and editing as a multi-turn process between agent and a CAD sandbox. At each turn, the agent perceives historical observations $\mathcal{O}_t$ to generate a structured action $a_t$ (reasoning and code block), receiving compiled visual or error feedback. Through two-stage optimization, the trained agent successfully acquires robust capabilities in multimodal engineering drawing parsing, precise parametric code synthesis, and long-horizon autonomous geometric self-correction.}
  \label{fig:pipeline}
\end{figure*}

\subsection{Task Formulation \& Interaction Protocol}
\label{subsec:problem_formulation}
We formulate interactive CAD generation as a multi-turn interaction between agent and execution sandbox, as shown in Fig.~\ref{fig:pipeline}. 
\method{} unifies three design modalities aiming to produce executable CadQuery programs: 
1) \emph{Drawing-to-Code}, conditioned on a multi-view engineering drawing $\mathcal{V}_{\text{ref}}$ with orthographic projections and dimensions; 
2) \emph{Text-to-Code}, guided by a natural language requirement $\mathcal{T}_{\text{ref}}$; 
3) \emph{Interactive Editing}, which performs localized modifications on a source program $\mathcal{C}_{\text{src}}$ based on incremental instructions. 
At each turn $t \le T$, the agent outputs a code variant $\mathcal{C}_t$, and the sandbox returns a multimodal feedback tuple $\mathcal{F}_t$. 
Crucially, beyond compiler logs, our sandbox leverages the OCCT kernel to programmatically project the generated solid into standard orthographic views and compute explicit dimensional annotations (linear extents and arc radii) directly from the underlying geometry as visual feedback $\mathcal{F}_{t}$. 
The session terminates upon emitting a completion token or reaching the maximum turn $T$.

\noindent \textbf{Observation \& Action Spaces.}
At turn $t$, the agent perceives the complete interaction history $\mathcal{O}_t = \{ \mathcal{S}, \mathcal{X}_{\text{ref}}, (\mathcal{C}_1, \mathcal{F}_1), \dots, (\mathcal{C}_{t-1}, \mathcal{F}_{t-1}) \}$, where $\mathcal{S}$ is the system prompt and $\mathcal{X}_{\text{ref}}$ is the initial reference context ($\mathcal{V}_{\text{ref}}$, $\mathcal{T}_{\text{ref}}$, or $\mathcal{C}_{\text{src}}$). 
This history preserves prior codes and multimodal feedback to empower evidence-based revisions. 
At each turn, the agent must output a structured action $a_t$ complying with a strict schema: 
1) a \textbf{Reasoning Trace} enclosed within \texttt{<thinking>} and \texttt{</thinking>} tags for multimodal feedback diagnosis and planning; and 
2) an \textbf{Execution Block} containing either a fenced CadQuery code block or a \texttt{<DONE>} token. 
Each turn must yield exactly one valid reasoning trace and one block; any violation triggers immediate trajectory invalidation and session termination.

\subsection{Training Dataset Curation}
\label{subsec:dataset}

\begin{figure*}[t]
\centering
\includegraphics[width=0.98\linewidth]{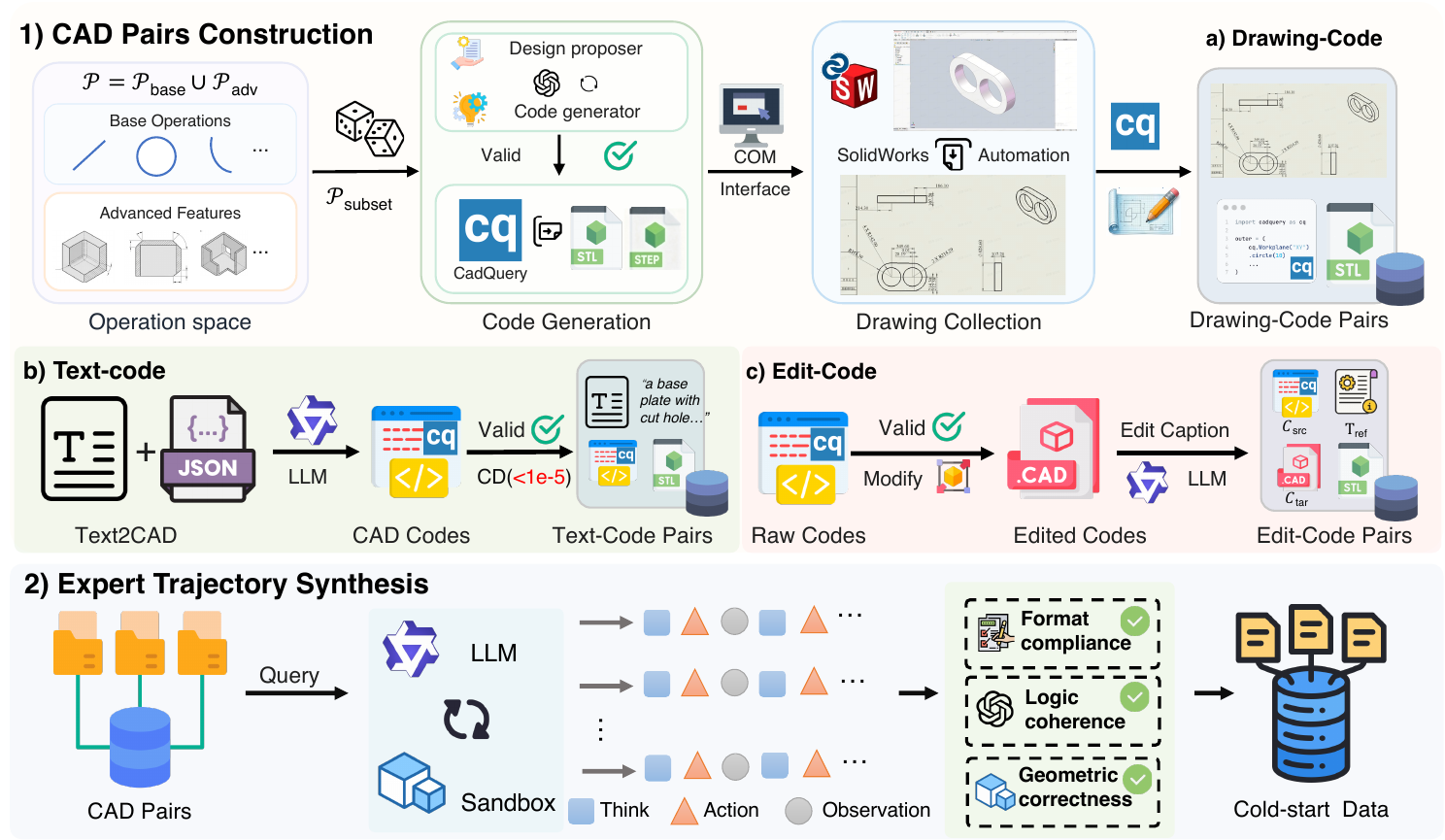}
\vspace{-1mm}
\caption{Data curation pipeline for \method{}. The pipeline first constructs three categories of high-quality CAD pairs. An expert LLM then rolls out multi-turn interactive trajectories within the CAD sandbox. Finally, the resulting cold-start corpus is distilled through format, logic, and geometric-correctness filters.}
\label{fig:data_pipeline}
\end{figure*}

We develop a specialized data curation pipeline for closed-loop CAD modeling as shown in Fig.~\ref{fig:data_pipeline}. Specifically, we synthesize high-quality training pairs across three heterogeneous configurations: \emph{Drawing-Code} pairs, \emph{Text-Code} pairs, and \emph{Edit-Code} pairs. Leveraging these foundational pairs, we then utilize an expert LLM to roll out multi-turn interaction trajectories, yielding a high-fidelity cold-start corpus for subsequent training.

\noindent \textbf{CAD Pairs Construction.}
To train and evaluate our framework across diverse design modalities, we construct three types of standardized datasets: 1) Drawing-Code Pairs, which overcome existing geometric limitations by defining an advanced operation space ($\mathcal{P}_{\text{base}}\cup\mathcal{P}_{\text{adv}}$) to synthesize complex parts, and leverage the SolidWorks Component Object Model (COM) interface to generate standard-compliant multi-view engineering drawings ($\mathcal{V}_{ref}$) with annotations; 2) Text-Code Pairs, which refine raw descriptions from Text2CAD~\cite{khan2024text2cad} into standardized texts ($T_{\text{ref}}$) and translate them into CadQuery code, strictly filtered via sandbox execution and Chamfer Distance; and 3) Edit-Code Pairs, which emulate realistic design revisions by applying controlled degradations to flawless scripts to form imperfect source code ($C_{\text{src}}$) paired with design-change instructions ($T_{\text{ref}}$) and target code ($C_{\text{tgt}}$). Detailed synthesis workflows, operation definitions, and filtering criteria are deferred to Appendix~\ref{app:cad_pair_construct}.

\noindent \textbf{Expert Trajectory Synthesis.}
For cold-start SFT, we synthesize an agentic trajectory dataset from the curated CAD pairs. For each instance, we prompt \texttt{Qwen3-VL-235B-A22B-Instruct}~\cite{bai2025qwen3} to roll out a multi-turn interactive trajectory under \method{}  protocol, formed by cascaded iterations of requirement analysis, executable CadQuery code, and sandbox feedback that repeat until the agent self-terminates via a \texttt{<DONE>} token. The resulting trajectories pass through three complementary checks: \emph{format compliance} discards outputs violating the prescribed reasoning-code schema; \emph{logic coherence} prompts an LLM to reject trajectories whose reasoning contradicts the executed actions or feedback; and \emph{geometric correctness} compiles each final program in the sandbox, retaining only trajectories that execute successfully and satisfy $CD < 10^{-5}$. Thus, we yield a high-fidelity cold-start corpus with both geometric accuracy and process-level supervision.

\subsection{Agent Training}
\label{subsec:training}

We train \method{} with a two-stage recipe. The first stage uses a progressive SFT to establish CAD syntax, drawing understanding, and basic editing behavior. The second stage applies reinforcement learning to optimize long-horizon refinement under executable and geometric feedback.

\noindent \textbf{Progressive Cold-start SFT.}
We initialize \method{} through a progressive supervised fine-tuning strategy built upon the multi-turn trajectory dataset introduced in Sec.~\ref{subsec:dataset}. We first perform cold-start SFT on expert trajectories to establish the fundamental capability of mapping engineering specifications to executable CadQuery programs. Formally, the training set is defined as $\mathcal{D}_{\text{S1}} = \{(x_i, y_i)\}$.

Building upon the base policy $\pi_{\theta}$ trained on $\mathcal{D}_{\text{S1}}$, we conduct an on-policy rollout stage to enhance interactive refinement. By executing $\pi_{\theta}$ inside the CAD sandbox, we collect trajectories that expose its typical execution failures and geometric discrepancies. A stronger teacher model then diagnoses these errors to synthesize corrective actions conditioned on environmental feedback and rendered observations, yielding the refinement dataset $\mathcal{D}_{\text{S2}}$. This curriculum enables the policy to explicitly learn iterative debugging and localized geometric refinement.
The overall supervised objective is optimized using the standard cross-entropy loss:
\begin{equation}
\mathcal{L}_{\text{SFT}}(\theta) = - \mathbb{E}_{(x_i, y_i) \sim (\mathcal{D}_{\text{S1}} \cup \mathcal{D}_{\text{S2}})} \left[ \log \pi_{\theta}(y_i \mid x_i) \right]
\end{equation}

\noindent \textbf{GSPO for Geometry-aware RL.}
Following the cold-start stage, we further optimize the policy via sequence-level reinforcement learning to enhance its closed-loop refinement capabilities. In our multi-turn CAD generation setting, each sampled trajectory contains interleaved reasoning traces, parametric code, and execution feedback across several turns. We therefore adopt GSPO~\cite{zheng2025group}, a sequence-level policy optimization method that estimates advantages within groups of sampled trajectories.

Specifically, given a design specification $x$, we sample a group of $G$
trajectories $\{y_g\}_{g=1}^G$ from the current policy
$\pi_{\theta_{\text{old}}}$ and evaluate their sequence-level rewards
$\{r_g\}_{g=1}^G$.
The relative advantages are computed via group-level standardization:
\begin{equation}
\hat{A}_{g} = \frac{r_{g} -
\mathrm{mean}(\{r_{g'}\}_{g'=1}^G)}{\mathrm{std}(\{r_{g'}\}_{g'=1}^G)}.
\end{equation}
Since trajectory lengths vary significantly across turns, we adopt a
length-normalized importance ratio at the sequence level to prevent
longer trajectories from dominating the gradient~\cite{chng2025sensenova}. The policy is then updated by maximizing the clipped surrogate objective over the training distribution $\mathcal{D}$:
{\small
\begin{equation}
J(\theta) = \mathbb{E}_{x \sim \mathcal{D}} \left[ \frac{1}{G} \sum_{g=1}^{G} \min\Big(\rho_g(\theta)\hat{A}_{g},\, \text{clip}_g(\theta)\hat{A}_{g}\Big) \right],
\end{equation}
}
where $\text{clip}_g(\theta) = \mathrm{clip}\big(\rho_g(\theta), 1-\epsilon_l, 1+\epsilon_h\big)$. $\rho_g(\theta)$ is the length-normalized importance ratio.
We adopt asymmetric clipping bounds $\epsilon_l < \epsilon_h$ to permit moderately larger updates along high-advantage directions while enforcing tighter constraints against policy degradation.

\noindent \textbf{Reward Modeling.}
To guide the RL optimization, we formulate the sequence-level reward for a trajectory $y$ as a
composite of three terms:
\begin{equation}
R(y) = R_{\mathrm{CD}}(y) + \lambda_f \, R_{\mathrm{fmt}}(y) +
\lambda_p \, R_{\mathrm{prog}}(y),
\end{equation}
where $\lambda_f$ and $\lambda_p$ scale the relative importance of compliance and progression, respectively.

\noindent\emph{Geometric reward.}
The primary feedback $R_{\mathrm{CD}} \in [0, 1]$ evaluates the shape fidelity via the Chamfer Distance ($CD$) between point clouds sampled from the generated and reference solids.
To ensure a variance-bounded signal, we map $CD$ into a bounded reward via a piecewise linear function:
{
\begin{equation}
R_{\mathrm{CD}}(y) =
\begin{cases}
1, & CD \le \delta_{\min} \\
\frac{\delta_{\max} - CD}{\delta_{\max} - \delta_{\min}}, & \delta_{\min} < CD < \delta_{\max} \\
0, & CD \ge \delta_{\max} \text{ or Fail}
\end{cases}
\end{equation}
}
where $\delta_{\min}$ and $\delta_{\max}$ define the strict compliance and filtering thresholds, respectively.

\noindent\emph{Format reward.}
The binary term $R_{\mathrm{fmt}} \in \{0, 1\}$ enforces strict adherence to our interaction protocol. It assigns $1$ only if every intermediate turn produces a structured reasoning trace paired with an executable code block, and the trajectory correctly concludes with an explicit completion marker.

\noindent\emph{Progress reward.}
To encourage substantive multi-turn refinement while preventing reward hacking, $R_{\mathrm{prog}}$ provides a sparse bonus if the trajectory satisfies three joint conditions: 1) spans at least two executable turns, 2) yields a valid geometry on the initial turn, and 3) achieves a monotonically decreasing $CD$ that converges below $\eta$. 

\noindent \textbf{Geometry-Viable Prefix Masking.}
While GSPO operates at the sequence level, multi-turn trajectories pose a distinct credit-assignment problem. Once the model emits code that triggers consecutive runtime errors or produces a geometry that ceases to improve, all subsequent turns inherit this corrupted context, making self-correction increasingly difficult. However,
their tokens still share the same sequence-level advantage $\hat{A}_g$, causing misattributed gradients that penalize viable early turns for downstream failures they did not cause.

To address this, GVPM monitors two complementary signals per trajectory to determine a prefix boundary $f = \min(f_{\mathrm{exec}}, f_{\mathrm{stall}})$:
1) \emph{Execution Cascade} ($f_{\mathrm{exec}}$) flags an irrecoverable failure loop indicated by $K$ consecutive runtime errors.
2) \emph{Geometry Stall} ($f_{\mathrm{stall}}$) detects when the Chamfer Distance across $K$ valid turns plateaus (fails to strictly decrease) while remaining above a quality gate $\eta$.
This boundary induces a per-token loss mask:
\begin{equation}
m_u = \mathbbm{1}\big[\tau(u) < f\big],
\end{equation}
where $\tau(u)$ maps token $u$ to its turn index, thereby excluding tokens with $m_u = 0$ from the policy loss. 
Furthermore, for masked trajectories, a one-sided advantage clamp $\hat{A}_g \leftarrow \max(\hat{A}_g, 0)$ is applied, ensuring that the viable prefix can still be reinforced but is never punished by downstream failures.

\subsection{New Benchmark and Metrics}
\label{subsec:benchmark}

To rigorously quantify the agent's performance in closed-loop CAD environments, we establish a comprehensive evaluation framework comprising two standardized task tracks and a suite of geometry-aware metrics.

\noindent \textbf{Evaluation Tasks.}
Our benchmark targets two core dimensions of interactive modeling. The first task, \textbf{\textit{IterCAD-Draw}}, evaluates the agent's capacity to reverse-engineer executable CadQuery code from dimensioned multi-view engineering drawings via closed-loop sandbox refinement. The second task, \textbf{\textit{IterCAD-Edit}}, presents a baseline script and tasks the agent with making localized, incremental modifications based on natural language instructions, thereby shifting the evaluative focus to parameter tuning and topology preservation.

\noindent \textbf{Unified Metrics.}
Standard CAD evaluation protocols typically rely on mean geometric distances, which suffer from survivor bias because non-executable samples ($CD \to \infty$) are omitted from the average. To provide a unified measurement of both code validity and geometric fidelity, we propose the \emph{Chamfer Distance Tolerance-Recall (CD-TR) Curve}. Formally, given a tolerance threshold $\tau$, the recall rate $R(\tau)$ is defined as the fraction of all test samples whose final generated geometry executes successfully and satisfies $CD \le \tau$, computed as: $R(\tau) = \frac{1}{M} \sum_{i=1}^{M} \mathbb{I}\big(\text{Valid}(y_i) \land CD(y_i) \le \tau\big)$, where $M$ is the total number of evaluation instances, $\mathbb{I}(\cdot)$ is the indicator function, and $\text{Valid}(y_i)$ denotes successful execution without geometric anomalies. By sweeping $\tau$ across a continuous spectrum of engineering tolerances, failed samples are naturally penalized as zero recall across all thresholds, eliminating survivor bias. In implementation, AUC-TR is integrated over a fixed log-scale CD tolerance range and normalized to $[0,1]$; detailed settings are provided in Appendix~\ref{app:metric_details}. To condense this profile into a single scalar for cross-model comparison, we report the \emph{Area Under the CD-TR Curve (AUC-TR)}. This joint index reflects both the programmatic robustness and geometric precision of the agent, serving as a rigorous evaluation standard for interactive CAD tasks.

\section{Experiments}

\begin{table*}[t]
\centering
\setlength{\tabcolsep}{5pt}
\renewcommand{\arraystretch}{1}
\small
\resizebox{\textwidth}{!}{
\begin{tabular}{l ccccc ccccc}
\toprule
 & \multicolumn{5}{c}{\textit{\textbf{Direct Inference}}} & \multicolumn{5}{c}{\textit{\textbf{Agentic Workflow}}} \\
\cmidrule(lr){2-6} \cmidrule(lr){7-11}
\textbf{Models} & IR\,(\%)\,$\downarrow$ & AUC-TR\,$\uparrow$ & Mean CD\,$\downarrow$ & Med. CD\,$\downarrow$ & Avg. Turn\,$\downarrow$ & IR\,(\%)\,$\downarrow$ & AUC-TR\,$\uparrow$ & Mean CD\,$\downarrow$ & Med. CD\,$\downarrow$ & Avg. Turn\,$\downarrow$ \\
\midrule
\multicolumn{11}{l}{\textit{Proprietary}} \\
GPT-5                  & 28.10 & 0.41 & 12.97 & 0.13 & 1.00 & 5.30  & 0.51 & 8.68 & 0.46 & 3.35 \\
Gemini-2.5-flash-lite  & 60.00 & 0.24 & 13.51 & 0.08 & 1.00  & 38.90 & 0.28 & 16.09 & 0.82 & 4.49 \\
Gemini-3-flash-preview    & 50.60 & 0.37 & \textbf{6.47}  & \textbf{0.06} & 1.00  & 12.60 & 0.55 & 7.03  & 0.10 & 4.31 \\
\cmidrule(lr){1-11}
\multicolumn{11}{l}{\textit{Open-source}} \\
GLM-4.6v               & 27.50 & 0.36 & 14.88 & 1.90 & 1.00  & 8.50  & 0.46 & 9.83 & 1.26 & 4.07 \\
InternVL3.5-30B-A3B    & 89.60 & 0.02 & 25.36 & 18.33 & 1.00  & 80.50 & 0.04 & 27.98 & 20.45 & 4.80 \\
InternVL3.5-8B         & 91.50 & 0.02 & 26.50 & 15.62 & 1.00  & 76.50 & 0.05 & 34.43 & 25.31 & 4.97 \\
Qwen3.5-35B-A3B        & 68.80 & 0.24 & 7.39  & \textbf{0.06} & 1.00  & 20.60 & 0.42 & 11.40 & 0.30 & 3.97 \\
Qwen3.5-4B & 95.30 & 0.03 & 9.22 & 0.07 & 1.00 & 64.60 & 0.24 & 10.48 & \textbf{0.07} & 4.56 \\
\cmidrule(lr){1-11}
\rowcolor{oursblue}
\textbf{\method{} (Ours)}  & \textbf{6.50} & \textbf{0.54} & 9.75 & 0.30 &  1.00  & \textbf{0.30} & \textbf{0.61} & \textbf{5.09} & 0.35 & \textbf{2.97} \\
$\Delta$ vs.\ Qwen3.5-4B & \imp{$-88.80$} & \imp{$+0.51$} & \reg{$+0.53$} & \reg{$+0.23$} & $0.00$ & \imp{$-64.30$} & \imp{$+0.37$} & \imp{$-5.39$} & \reg{$+0.28$} & \imp{$-1.59$} \\
\bottomrule
\end{tabular}
}
\caption{Performance on \textbf{\textit{IterCAD-Draw}} bench. Comparison between Direct Inference and Agentic Workflow.}
\label{tab:codepass_compact_horizontal}
\end{table*}
 
\subsection{Implementation Details}
\label{subsec:implementation}

\noindent \textbf{Datasets \& Infrastructure.} Our training corpus comprises $28$K SFT trajectories ($20$K expert $\mathcal{D}_{\text{S1}}$ for code generation, $8$K on-policy $\mathcal{D}_{\text{S2}}$ for iterative refinement) and $2$K hard Drawing-to-Code samples for RL. Utilizing Swift~\cite{zhao2025swift} on 8 NVIDIA A100 GPUs, we initialize from \texttt{Qwen3.5-4B}~\cite{qwen35blog}. SFT runs for 3 epochs at a learning rate (LR) of $1\times10^{-5}$. The subsequent GSPO RL stage uses a group size $G=8$ and LR of $1\times10^{-6}$ over a maximum of $T=5$ interaction turns. Reward weights are set to $\lambda_f=0.5$, and $\lambda_p=0.2$. For GVPM, the trigger window is $K=2$. Fully expanded implementation setups are detailed in Appendix~\ref{app:implementation_details}.

\begin{wrapfigure}[17]{r}{0.5\linewidth}
\centering
\includegraphics[width=\linewidth]{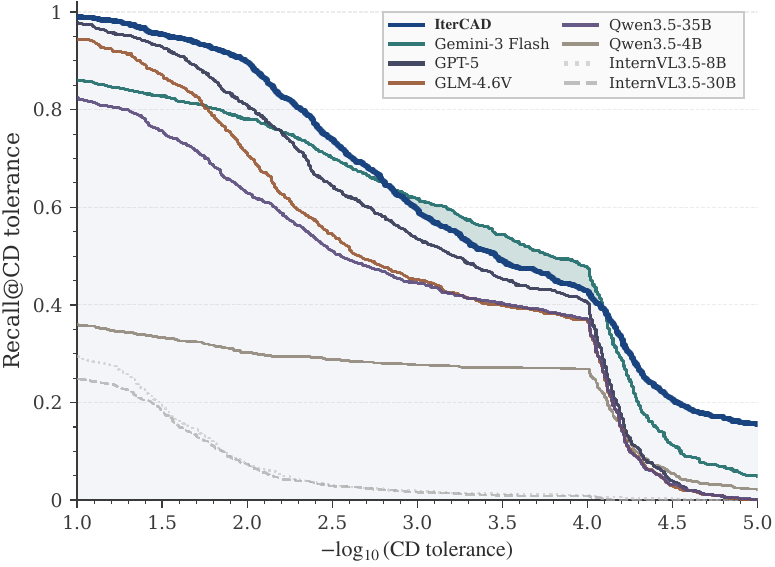}
\vspace{-10pt}
\caption{CD-TR Curve on \textbf{\textit{IterCAD-Draw}} bench.}
\label{fig:cd-curve}
\vspace{-10pt}
\end{wrapfigure}

\noindent \textbf{Benchmarks.} We evaluate multi-turn generation across: 1) \method{}-Bench: Our proposed suite with $1$K drawing and $200$ editing tasks; 2) Text2CAD Bench~\cite{khan2024text2cad}: $8,046$ multimodal parts with text specifications; and 3) CADPrompt Bench~\cite{alrashedy2025generating}: $200$ expert instructions for zero-shot text-to-CAD synthesis.

\noindent \textbf{Evaluation Metrics.} Performance is assessed via a multi-dimensional metric suite: Invalid Ratio (IR): the proportion of programs triggering syntax exceptions or geometric anomalies; Mean/Median Chamfer Distance (Mean/Med. CD): geometric fidelity computed over successful executions, with all reported CD values multiplied by $10^3$; Average Turn (Avg. Turn): multi-turn interactive efficiency. Crucially, to eliminate survivor bias, we report the CD-TR Curve and its scalar counterpart, AUC-TR, as the definitive indicators of joint execution robustness and geometric convergence.

\begin{table}[t]
\centering
\setlength{\tabcolsep}{4pt}
\renewcommand{\arraystretch}{0.8}
\small
\resizebox{0.75\columnwidth}{!}{%
\begin{tabular}{l ccccc}
\toprule
\textbf{Model} & IR\,(\%)\,$\downarrow$ & AUC-TR\,$\uparrow$ & Mean CD\,$\downarrow$ & Med. CD\,$\downarrow$ & Avg. Turn\,$\downarrow$ \\
\midrule
\multicolumn{6}{l}{\textit{Proprietary}} \\
GPT-5                  & \textbf{0.50} & \textbf{0.79} & \textbf{2.14} & \textbf{0.05} & \textbf{2.31} \\
Gemini-2.5-flash-lite  & 43.00 & 0.34 & 8.55  & 0.31 & 4.38 \\
Gemini-3-flash-preview    & 11.00 & 0.51 & 4.27  & 0.06 & 4.23 \\
\cmidrule(lr){1-6}
\multicolumn{6}{l}{\textit{Open-source}} \\
GLM-4.6v               & 14.00 & 0.44 & 10.69 & 1.49 & 3.74 \\
InternVL3.5-30B-A3B    & 67.50 & 0.16 & 8.93  & 2.18 & 4.48 \\
InternVL3.5-8B         & 65.50 & 0.17 & 10.44 & 1.91 & 4.36 \\
Qwen3.5-35B-A3B        & 11.00 & 0.55 & 7.49  & 0.14 & 3.84 \\
Qwen3.5-4B  & 63.00 & 0.18 & 9.96 & 1.25 & 4.49 \\
\cmidrule(lr){1-6}
\rowcolor{oursblue}
\textbf{\method{} (Ours)}  & 1.00 & 0.54 & 7.52 & 0.77 & 2.34 \\
$\Delta$ vs.\ Qwen3.5-4B & \imp{$-62.00$} & \imp{$+0.36$} & \imp{$-2.44$} & \imp{$-0.48$} & \imp{$-2.15$} \\
\bottomrule
\end{tabular}
}
\vspace{10pt}
\caption{Performance on the \textbf{\textit{IterCAD-Edit}} bench.}
\vspace{-10pt}
\label{tab:edit_codepass}
\end{table}

\subsection{Main Results}

\noindent \textbf{Interactive Drawing-to-Code Generation.}
Tab.~\ref{tab:codepass_compact_horizontal} reports the comprehensive evaluation of \method{} on the \textit{IterCAD-Draw} task. Under the \emph{Direct Inference} setting, \method{} attains the lowest invalid ratio (IR) of $6.50\%$ and the highest AUC-TR of $0.54$, decisively surpassing both proprietary leaders like GPT-5 ($28.10\%$ IR) and open-source baselines like GLM-4.6v ($27.50\%$ IR); notably, relative to its \texttt{Qwen3.5-4B} backbone, it reduces IR by $88.80$ points and lifts AUC-TR by $0.51$, proving our training recipe instills robust single-pass code validity prior to interactive refinement. This advantage becomes more pronounced under the \emph{Agentic Workflow} using closed-loop sandbox feedback: \method{} efficiently converts the multi-turn budget into superior accuracy, achieving a near-zero IR of $0.30\%$, the best AUC-TR of $0.61$, and the lowest mean Chamfer Distance of $5.09$ within only $2.97$ average turns. Notably, this low average turn implies that most instances are successfully generated in a single pass, utilizing the subsequent turn primarily for validation rather than extensive revision. In contrast, the \texttt{Qwen3.5-4B} baseline consumes $4.56$ turns yet still suffers from a $64.60\%$ IR, while much larger competitors (e.g., Qwen3.5-35B-A3B and InternVL3.5-30B-A3B) remain far behind in joint robustness and fidelity, confirming that \method{} performs autonomous error correction more reliably and efficiently than models an order of magnitude larger.

\begin{wraptable}[10]{!t}{0.5\linewidth}
\centering
\vspace{-10pt}
\setlength{\tabcolsep}{3pt}
\renewcommand{\arraystretch}{0.85}
\begin{tabular}{lccc}
\toprule
Method & IR\% $\downarrow$ & Mean CD $\downarrow$ & Med. CD $\downarrow$ \\
\midrule
Claude-3.7-sonnet & 47.03         & 186.53          & 134.16        \\
GPT-4o            & 93.00         & 133.52          & 45.91         \\
Deepseek-V3       & 51.96         & 186.69          & 107.57        \\
Qwen2.5-72B       & 82.64         & 209.41          & 153.81        \\
Qwen2.5-7B        & 98.83         & 202.35          & 169.86        \\
Text2CAD          & 3.75          & 29.29           & 0.37          \\
CAD-Coder         & 1.45          & \textbf{6.54}   & 0.17          \\
\rowcolor{oursblue}
\textbf{\method{}} & \textbf{0.64} & 10.92           & \textbf{0.10} \\
\bottomrule
\end{tabular}
\vspace{-4pt}
\caption{Evaluation results on the Text2CAD bench.}
\label{tab:text2cad_res}
\end{wraptable}

\noindent \textbf{Instruction-based Editing.}
As summarized in Tab.~\ref{tab:edit_codepass}, \method{} substantially outperforms its \texttt{Qwen3.5-4B} backbone, reducing IR by $62.00\%$ and shortening convergence by $2.15$ turns. While GPT-5 leads in AUC-TR due to its frontier scaling advantage in text-to-code reasoning and complex program refactoring, \method{} achieves an exceptionally low IR ($1.00\%$) and a comparable efficiency ($2.34$ turns) that nearly match GPT-5. This validates that our interactive learning paradigm effectively teaches the agent precise code editing without relying on massive parameter scales.

\begin{wraptable}[12]{!t}{0.5\linewidth}
\centering
\setlength{\tabcolsep}{3pt}
\renewcommand{\arraystretch}{0.85}
\begin{tabular}{lccc}
\toprule
Method & IR\% $\downarrow$ & Mean CD $\downarrow$ & Med. CD $\downarrow$ \\
\midrule
Text2CAD      & 6.00 (1.57) & -             & 127.70        \\
CADmium       & 10.50 (6.28)          & -             & 116.75        \\
CAD-Judge     & 4.50 (2.55)          & 152.40        & 42.69         \\
\rowcolor{oursblue}
\textbf{\method{}} & \textbf{2.00}          & \textbf{10.45} & \textbf{2.42}  \\
\bottomrule
\end{tabular}
\caption{Evaluation results on the CADPrompt bench. Parentheses show CAD-Judge's~\cite{zhou2026cad} original metrics; we re-evaluate IRs due to calculation inconsistencies on the 200-sample set.}
\label{tab:cadprompt_results}
\end{wraptable}

\noindent \textbf{Text-to-Code Benchmarks.}
We further validate \method{} on two widely adopted external text-driven CAD benchmarks. On Text2CAD (Tab.~\ref{tab:text2cad_res}), \method{} achieves an exceptionally low IR of $0.64\%$ and a Med. CD of $0.10$, heavily outperforming general-purpose models like GPT-4o and DeepSeek-V3, which suffer from IRs over $50\%$. Compared to the domain-specific CAD-Coder, \method{} strikes a better balance between code validity and median geometric precision. On the challenging CADPrompt benchmark (Tab.~\ref{tab:cadprompt_results}), \method{} again establishes prominent advantages in geometric fidelity, yielding a Mean CD of $10.45$ and a Med. CD of $2.42$ (outperforming CAD-Judge by an order of magnitude) with only a $2\%$ IR. These consistent advantages confirm the strong capabilities of our closed-loop architecture under diverse evaluation protocols.

\subsection{Ablation Study}

\begin{table}[bp]
\centering
\setlength{\tabcolsep}{4pt}
\renewcommand{\arraystretch}{1}
\small
\resizebox{0.9\columnwidth}{!}{%
\begin{tabular}{lcccccccc}
\toprule
\multirow{2}{*}{\textbf{Variant}} & \multicolumn{4}{c}{\textbf{Training Component}} & \multicolumn{4}{c}{\textbf{Metric}} \\
\cmidrule(lr){2-5} \cmidrule(lr){6-9}
& ${\mathcal{D}_{\text{S1}}}$  & ${\mathcal{D}_{\text{S1}} \cup \mathcal{D}_{\text{S2}}}$ & GSPO & GVPM & IR$\downarrow$ & AUC-TR$\uparrow$ & Mean CD$\downarrow$ & Avg. Turn$\downarrow$ \\
\midrule
Qwen3.5-4B & -- & -- & -- & -- & 64.60 & 0.24 & 10.48 & 4.56 \\
\method{}-$\mathrm{SFT}_1$ & \imp{$\checkmark$} & -- & -- & -- & 7.50 & 0.52 & 10.55 & 2.23 \\
\method{}-$\mathrm{SFT}_2$ & \imp{$\checkmark$} & \imp{$\checkmark$} & -- & -- & 0.80 & 0.52 & 12.44 & 3.08 \\
\method{}-GSPO & \imp{$\checkmark$} & \imp{$\checkmark$} & \imp{$\checkmark$} & -- & 1.30 & 0.58 & 8.00 & \textbf{2.51} \\
\rowcolor{oursblue}
\textbf{\method{}-Full} & \imp{$\checkmark$} & \imp{$\checkmark$} & \imp{$\checkmark$} & \imp{$\checkmark$} & \textbf{0.30} & \textbf{0.61} & \textbf{5.09} & 2.97 \\
\bottomrule
\end{tabular}
}
\vspace{10pt}
\caption{Ablation study on \textbf{IterCAD-Draw} bench.}
\label{tab:ablation_components}
\end{table}

We conduct a comprehensive ablation on the \textit{IterCAD-Draw} task in Tab.~\ref{tab:ablation_components} to isolate each component's contribution. Starting from the \texttt{Qwen3.5-4B} backbone, the first cold-start SFT on expert trajectories ($\mathcal{D}_{\text{S1}}$) establishes the basic mapping from specifications to programs, reducing the invalid ratio (IR) from $64.60\%$ to $7.50\%$ and raising AUC-TR from $0.24$ to $0.52$. Augmenting training with on-policy refinement data ($\mathcal{D}_{\text{S1}} \cup \mathcal{D}_{\text{S2}}$) further drives IR down to $0.80\%$, demonstrating that teacher-corrected trajectories successfully instill iterative debugging behaviors.
However, SFT alone leaves Mean CD relatively high ($12.44$), proving imitation insufficient for fine-grained geometric optimization. Introducing GSPO RL addresses this by lowering Mean CD to $8.00$ and lifting AUC-TR to $0.58$, validating that closed-loop reinforcement achieves geometric refinement beyond simple imitation. While GSPO alone reduces average turns to $2.51$ via premature termination, integrating GVPM yields our full model and achieves the best overall performance: IR drops to $0.30\%$, AUC-TR reaches $0.61$, and Mean CD falls sharply to $5.09$. By masking tokens beyond the viable prefix, GVPM mitigates the multi-turn credit-assignment problem, reinforcing productive early actions rather than penalizing them for downstream failures, which safely shifts the agent toward substantive refinement despite a modest increase in average turns. Additional difficulty-stratified results, visual-feedback ablations, and Hard-subset failure analysis are provided in Appendix~\ref{app:additional_exp_analysis}.

\subsection{Case Study}

Beyond quantitative analysis, we present \method{} inference trajectories for all three tasks in Appendix~\ref{app:case_study}. Notably, Fig.~\ref{fig:CAD-gen_edit} in Appendix~\ref{app:case_study} shows a unified session where the \method{} reconstructs a base plate from a drawing and then applies successive natural-language edits, preserving unaffected geometry and resolving historical references such as ``undo the previous operation,'' demonstrating faithful, controllable CAD evolution process.

\section{Conclusion}

In this work, we presented \method{}, a unified, drawing-driven CAD agent framework that transforms from-scratch synthesis and instruction-based editing into a closed-loop interaction with an execution sandbox. By leveraging multi-view engineering drawings with dimensional constraints, \method{} provides precise physical anchors for iterative defect diagnosis, overcoming the limitations of syntax- or point-cloud-based feedback. Equipped with progressive SFT, geometry-aware RL, and Geometry-Viable Prefix Masking, our agent significantly improves code executability and geometric convergence. Comprehensive evaluations using holistic metrics (CD-TR Curve and AUC-TR) demonstrate that \method{} sets a more rigorous standard for robust and faithful CAD automation.
\section*{Limitations}

Despite these promising results, several limitations remain to be addressed in future work. 
First, our research focuses primarily on single-part parametric modeling within the CadQuery program and specific drawing standards, representing an initial exploratory step toward automated engineering drawing reverse engineering. Transitioning to other proprietary drawing specifications and large-scale assembly-level collaborative design is a crucial next step in validating the scalability of the entire ecosystem.
Second, although our benchmark incorporates advanced modeling features, it largely relies on geometric metrics like chamfer distance that fail to fully capture high-level mechanical semantics and functional manufacturing intents. Consequently, while the synthesized models may satisfy basic spatial dimensions, \method{} cannot yet recognize the underlying engineering utilities of specific features, such as keyways, bearing fits, or thread specifications. 
Finally, while the agent reliably delivers executable code, it prioritizes immediate geometric convergence over long-term program maintainability. This occasionally results in suboptimal, hard-coded implementations that lack the structured, hierarchical parametric design trees essential for human collaboration and engineering maintenance.

\section*{Acknowledgments}

This work is supported by Shanghai Artificial Intelligence Laboratory.

\clearpage
{
\bibliographystyle{unsrtnat}  
\bibliography{preprint}
}

\clearpage
\newgeometry{
  textheight=9in, textwidth=5.5in, top=1in,
  headheight=12pt, headsep=25pt, footskip=30pt
}

\cleardoublepage
\appendix
\noindent{\LARGE\textbf{Appendix}\par}\normalsize

\vspace{10pt}
{\large \textbf{Contents}}

\startcontents[appendices]
\printcontents[appendices]{l}{1}{\setcounter{tocdepth}{3}}

\section{Related Work}
\label{app:related_work}

\subsection{CAD Representations and Generation}
Learning-based 3D CAD generation fundamentally relies on underlying data representations, which govern how spatial geometry and design intents are modeled, which can be categorized into five paradigms: 
1) \textit{CSG-based methods}~\cite{du2018inversecsg, yu2022capri} generate shapes via Boolean operations but fail at representing intricate free-form surfaces; 
2) \textit{B-rep methods}~\cite{jayaraman2022solidgen, xu2024brepgen} directly predict static boundary networks but suffer from severe non-parametric outputs and high risks of topological violations; 
3) \textit{Sequence-based methods}, pioneered by DeepCAD~\cite{wu2021deepcad} and extended by Text2CAD~\cite{khan2024text2cad}, encode structures as tokenized command lines (e.g., \texttt{[SOL]}) to minimize inference sequences, yet lack standard runtime support for cross-entity revisions; 
4) \textit{Structured-text paradigms}, including CAD-Llama's SPCC~\cite{li2025cad} and Seek-CAD's SSR~\cite{li2025seek}, utilize customized, lightweight syntax to bypass massive software dependencies but face limited industrial ecosystem integration; and 
5) \textit{Code-based frameworks} (e.g., CAD-RL~\cite{niu2026intent}, CAD-Coder~\cite{guan2026cad}, CADrille~\cite{kolodiazhnyi2025cadrille}) generate fully parametric, executable scripts (e.g., CadQuery), leveraging robust geometric engines to maximize editing flexibility and LLM reasoning compatibility. 
Despite these programmatic advantages, existing generation frameworks remain fundamentally constrained by a \textit{one-shot open-loop} paradigm that lacks iterative self-correction. More critically, existing synthetic benchmarks~\cite{willis2021fusion, khan2024text2cad} are heavily dominated by trivial sketch-and-extrude loops, leaving advanced engineering features such as fillets, chamfers, and shells severely underrepresented.

\subsection{Multi-Turn CAD Agents}
To transition from open-loop generation to interactive refinement, recent agentic systems have coupled LLMs with execution environments to iteratively repair design defects. For instance, CAD-Editor~\cite{yuan2025cad} and PR-CAD~\cite{an2026pr} explore text-guided parametric script editing and progressive sequence infilling, while systems like CADDesigner~\cite{fan2025caddesigner} and TOOLCAD~\cite{gong2026toolcad} leverage multimodal dialogue and reinforcement learning for interactive 3D modeling. To guide these multi-turn updates, current methods primarily rely on two types of feedback signals. The first is environment-based execution and compiler feedback~\cite{zhou2026cad}, which verifies syntax validity but remains completely blind to semantic and geometric errors as long as the code executes without crashing. The second is distance-based spatial feedback~\cite{kabisov2026cadreasoner}, which computes discrepancies like Chamfer Distance between abstract 3D point clouds or rendered images to guide script rewriting.
However, the supervisory feedback leveraged by these existing multi-turn systems remains critically coarse and ambiguous for precise design orchestration. Text-based instructions lack structural anchors, while point-cloud or visual discrepancies only capture global, unstructured shape errors. Such spatial deviations fail to localize errors or attribute them to specific 2D sketch dimensions, geometric constraints, feature annotations, or local topological entities, frequently causing multi-turn iterations to degenerate into a directionless and unguided blind search. 

\section{CAD Pairs Construction}
\label{app:cad_pair_construct}

\subsection{Drawing-Code Pairs.}
Widely used datasets, such as DeepCAD~\cite{wu2021deepcad}, are strictly confined to basic "sketch-and-extrude" workflows and completely lack critical manufacturing features like fillets and chamfers, which severely limits their ability to represent the geometric complexity and engineering intent of real-world parts. To bridge this gap, we formally define an operation space $\mathcal{P}=\mathcal{P}_{\text{base}}\cup\mathcal{P}_{\text{adv}}$. The base set $\mathcal{P}_{\text{base}}$ comprises sketch primitives and fundamental extrusion constraints, while the advanced set $\mathcal{P}_{\text{adv}}$ encapsulates professional features such as fillets, chamfers, shells, and pattern arrays.
To synthesize highly complex geometries, our engine randomly samples a subset of heterogeneous operations $\mathcal{P}_{\text{sub}} \subset \mathcal{P}$ as a strict execution constraint for the GPT-5~\cite{openai2025gpt5}, forcing it to compose intricate CAD plans using the designated combinations. The output CadQuery code is then evaluated in a headless sandbox to verify syntax correctness and geometric validity (e.g., eliminating self-intersections). Only error-free programs are retained and exported as STL and STEP files.

Crucially, in order to establish standard-compliant visual supervision, we leverage the SolidWorks Component Object Model (COM) interface to programmatically convert 3D STEP files into multi-view engineering drawings $\mathcal{V}_{ref}$ featuring orthographic projections and explicit dimensional annotations. Rather than relying on simple rendered images, this mechanism provides reliable physical anchors for dimensional reasoning. After applying automated quality checks to filter out invalid geometry, non-manifold structures, and inconsistent annotations, we curate the final dataset as a collection of paired engineering drawings $\mathcal{V}_{ref}$ and executable CadQuery code $\mathcal{C}_{tgt}$.

\subsection{Text-Code Pairs.}
We build text-code pairs upon the Text2CAD~\cite{khan2024text2cad} dataset, which provides raw expert descriptions specifying the geometric and functional requirements of each part. To ensure instruction quality, we prompt an LLM to refine and reorganize these raw descriptions into standardized reference texts $T_{\text{ref}}$. We then employ an LLM to translate each $T_{\text{ref}}$, along with its structured parametric annotations, into CadQuery programs. Each program is compiled in a headless sandbox to verify syntactic correctness, and its geometric fidelity is assessed via Chamfer Distance ($CD$). We retain only programs that execute successfully and satisfy $CD < 10^{-5}$ as the target code $C_{\text{tgt}}$, yielding final text-code pairs $(T_{\text{ref}}, C_{\text{tgt}})$ that faithfully ground high-quality descriptions in geometrically accurate CAD code.

\subsection{Edit-Code Pairs.}
Beyond from-scratch generation, we synthesize code-rectification tasks emulating realistic design revisions. Starting from a validated CAD script, we apply controlled degradations—such as parametric perturbations, misplaced vertices, and feature substitutions—to construct an imperfect source program $C_{\text{src}}$. For each corrupted instance, we pair it with a concise design-change instruction $T_{\text{ref}}$, while treating the original flawless program as the target $C_{\text{tgt}}$. This protocol generates triplets $(C_{\text{src}}, T_{\text{ref}}, C_{\text{tgt}})$ that rigorously evaluate the agent's ability to repair specified geometric discrepancies while preserving unaffected code structures.

\section{Extended Implementation Details and Evaluation Protocols}
\label{app:implementation_details}

\begin{figure*}[!t]                                                           
  \centering
  \includegraphics[width=0.9\linewidth]{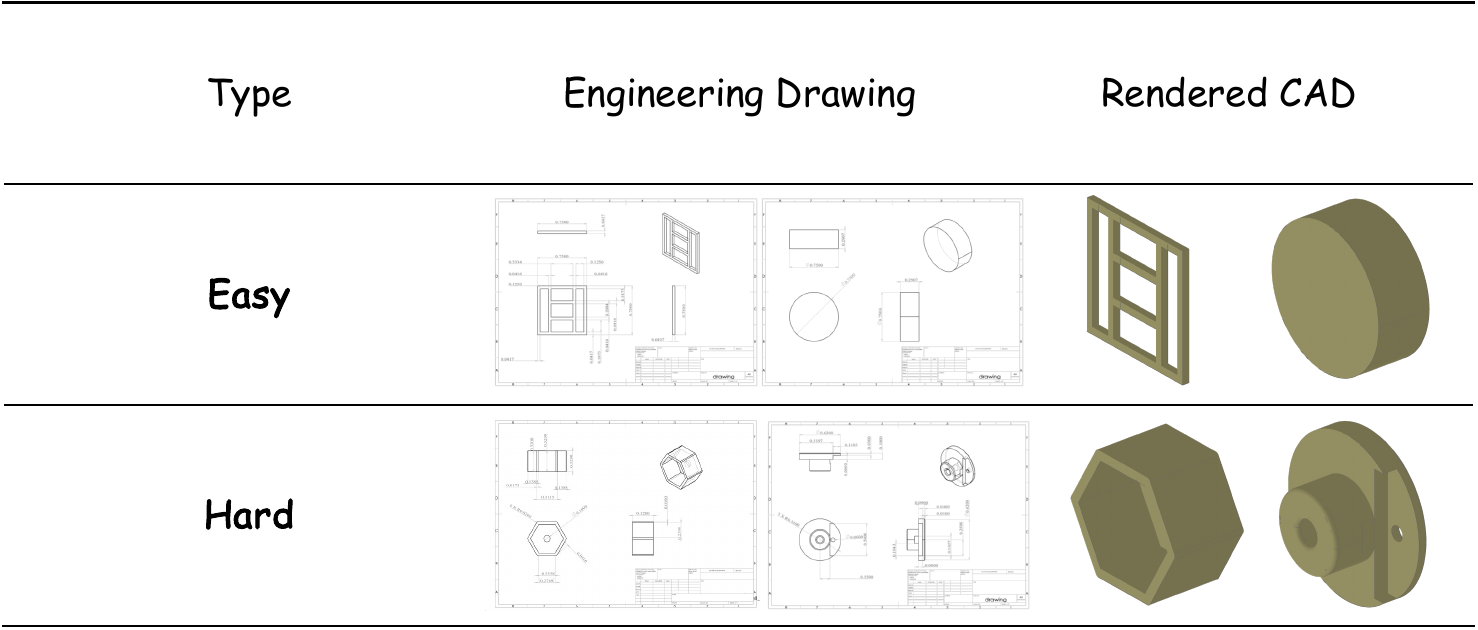}             
  \caption{Representative samples from the \textit{IterCAD-Draw} benchmark
across two difficulty levels, showcasing multi-view engineering drawings   
paired with ground-truth 3D geometries. Complexity increases from simple
extruded profiles (Easy-level) to parts requiring advanced operations such as   
shells, fillets, and through-cuts (Hard-level).}
  \label{fig:IterCAD-Draw}
  \vspace{-10pt}                                                             
\end{figure*}

\subsection{Dataset Curation and Training Budgets}
Following the curation pipeline in Sec.~\ref{subsec:dataset}, our training corpus spans three heterogeneous task types: \emph{Drawing-Code}, \emph{Text-Code}, and \emph{Edit-Code} pairs. For progressive cold-start SFT, we use a total of \textbf{28K} trajectories: \textbf{20K} expert trajectories for phase-1 ($\mathcal{D}_{\text{S1}}$), which establishes the basic mapping from engineering specifications to executable CadQuery programs, and \textbf{8K} on-policy refinement trajectories for phase-2 ($\mathcal{D}_{\text{S2}}$), synthesized by a teacher model that corrects the phase-1 policy's execution failures and geometric discrepancies to instill iterative refinement. For the RL stage, we restrict optimization to \textbf{2K} hard \emph{Drawing-to-Code} samples, as this task poses the most demanding multi-view reasoning and self-correction challenge and thus benefits most from closed-loop geometric feedback.
All \textit{IterCAD-Draw} ($1{,}000$) and \textit{IterCAD-Edit} ($200$) test instances are strictly excluded from training. Before training we run a scripted cross-split geometric check for exact overlaps and near-duplicates: none are found, and the minimum normalized Chamfer Distance between any training and test sample is $1.10$. Cold-start trajectories are retained only after format compliance, GPT-5 logic-coherence filtering (rejecting reasoning that contradicts executed actions or feedback), and geometric correctness ($CD<10^{-5}$). A manual spot-check of sampled accepted trajectories finds consistently high instruction fidelity, feedback-consistent correction, and final geometric agreement.

\subsection{Detailed Infrastructure and Hyperparameters}
\method{} is initialized from the \texttt{Qwen3.5-4B} backbone and trained utilizing the Swift~\cite{zhao2025swift} framework on 8 NVIDIA A100 GPUs. 
During the SFT stage, we perform autoregressive fine-tuning on the curated trajectories ($\mathcal{D}_{\text{S1}} \cup \mathcal{D}_{\text{S2}}$) with a learning rate of $1\times10^{-5}$ for 3 epochs. 
In the subsequent RL stage, we deploy GSPO to refine the policy's closed-loop self-correction capability, configuring the optimization with a group size of $G=8$, and a learning rate of $1\times10^{-6}$. 
Each training trajectory permits up to $T=5$ interaction turns with the CAD sandbox, during which the agent iteratively revises its code under execution and rendered feedback. 
The composite reward function defined in Sec.~\ref{subsec:training} balances geometric fidelity, format compliance, and multi-turn progression, with the corresponding weights empirically set to $\lambda_f=0.5$, and $\lambda_p=0.2$.
For GVPM, we set the trigger window to $K=2$, where the masking mechanism activates once a trajectory exhibits either two consecutive runtime errors or two consecutive valid-but-stalled turns above the quality gate.

\noindent \textbf{Agentic evaluation protocol.}
Every compared model, including proprietary baselines, is evaluated under an identical interaction protocol: the same system prompt (Fig.~\ref{fig:cad_system_prompt}), the same CAD sandbox, the same feedback content and format, a maximum of $T{=}5$ turns, and the same metric evaluator. A session terminates only when the model emits \texttt{<DONE>} or reaches the turn limit. The only varying component is the underlying MLLM.

\begin{figure*}[!t]
  \centering
  \includegraphics[width=0.9\linewidth]{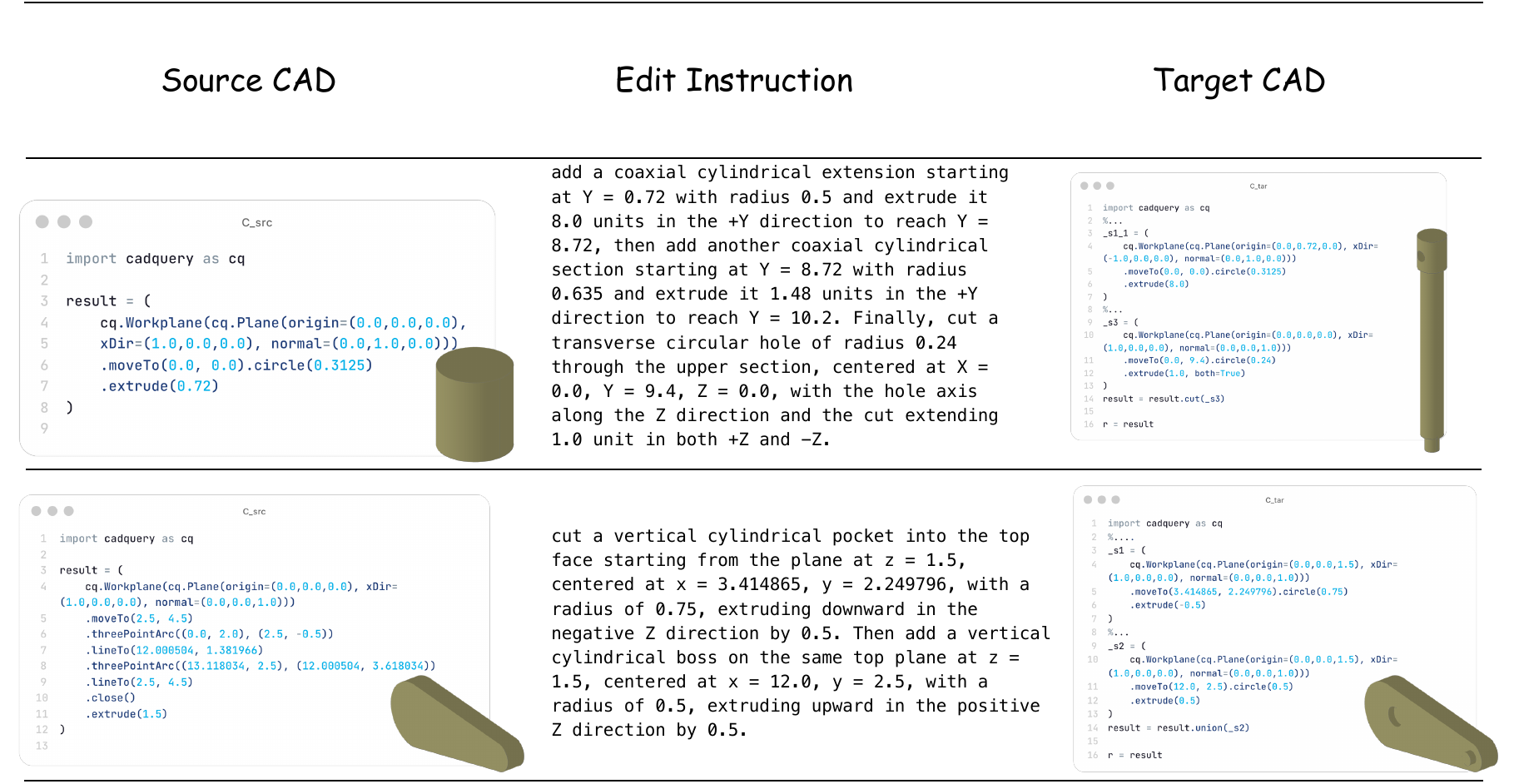}
  \caption{Representative samples from the \textit{IterCAD-Edit} benchmark.  
Each pair shows the source code (left), the editing     
instruction (middle), and the target code after modification (right),    
illustrating diverse edit operations including feature addition, Boolean     
subtraction, and parametric adjustment.}
  \label{fig:IterCAD-Edit}
\end{figure*}

\subsection{Benchmark Specifications}
To rigorously assess the performance of the CAD agent under multi-round, interactive settings, we evaluate across three complementary diagnostic suites:\method{}-Bench (Ours): An interactive evaluation suite comprising $1$K drawing-to-code tasks and $200$ local editing tasks. It challenges models to reverse-engineer standard-compliant engineering drawings into constructive code or interpret natural language revision instructions. Text2CAD Bench~\cite{khan2024text2cad}: A large-scale public baseline consisting of $8,046$ multimodal parts paired with natural language text specifications and oracle analytical geometries. CADPrompt Bench~\cite{alrashedy2025generating}: A curated zero-shot evaluation suite containing $200$ expert-level instructions designed to stress-test text-to-CAD synthesis across edge-case geometries.  
Figs.~\ref{fig:IterCAD-Draw} and~\ref{fig:IterCAD-Edit} provide representative examples of our drawing-driven generation and instruction-based editing tracks, respectively.

\begin{figure*}[!t]
  \centering                                                          
  \includegraphics[width=0.98\linewidth]{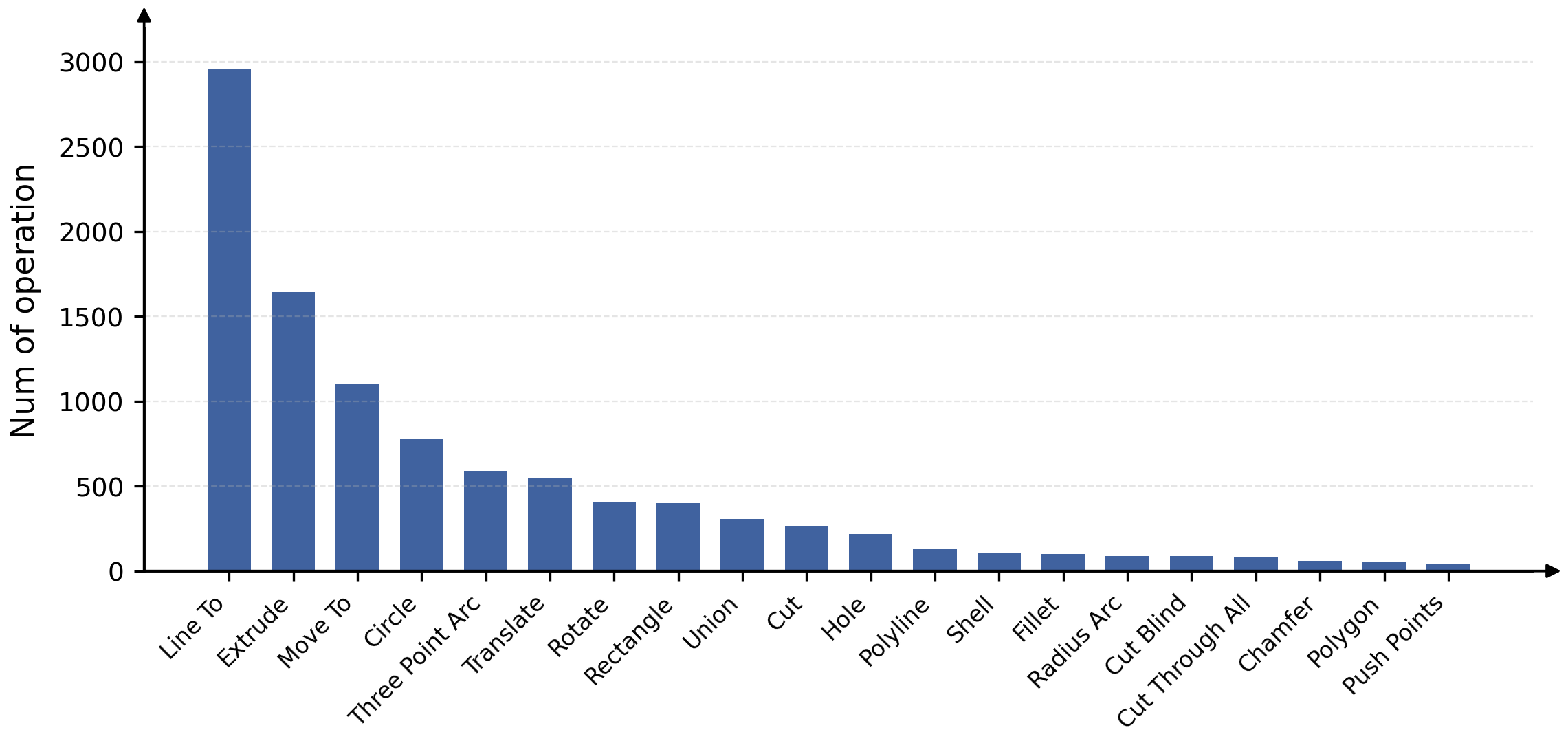}
  \caption{Top-20 CadQuery API operation distribution in the
\textit{IterCAD-Draw} benchmark. The histogram reveals a long-tail           
distribution where foundational operations (e.g., \texttt{extrude},
\texttt{rect}, \texttt{circle}) dominate, while advanced operations (e.g.,   
\texttt{shell}, \texttt{fillet}, \texttt{chamfer}) appear increasingly in
higher difficulty levels.}
  \label{fig:benchmark-status}
  \vspace{-15pt}
\end{figure*}

\subsection{IterCAD-Bench Details}

We provide additional details on the construction and composition of our     
evaluation benchmark, which comprises two complementary tasks:         
\textit{IterCAD-Draw} and \textit{IterCAD-Edit}.

\noindent \textbf{IterCAD-Draw.}                                             
The drawing-to-code track contains $1{,}000$ test instances stratified into
2 difficulty levels based on geometric and operational complexity:       
\begin{itemize}[leftmargin=*, itemindent=0pt, itemsep=0pt, topsep=2pt]                                                  
    \item \textbf{Easy} ($500$ samples): Parts primarily composed of      
sketch-and-extrude sequences with simple planar profiles (rectangles,        
circles, polygons). These samples test the agent's basic ability to parse
multi-view projections and synthesize valid CadQuery programs.               
    \item \textbf{Hard} ($500$ samples): Parts involving moderately
complex operations such as holes, blind cuts, and Boolean combinations atop extruded bodies, as well as parts requiring advanced
manufacturing operations including shell, fillet, chamfer, through-cuts, and their compositions. This tier evaluates the agent's capacity to handle
intricate topological modifications and multi-operation chaining.            
\end{itemize}

Each sample is paired with a ground-truth CadQuery program, a normalized STL
mesh, a STEP solid, and a standard-compliant multi-view engineering drawing  
rendered via the SolidWorks COM interface. Fig.~\ref{fig:benchmark-status}
visualizes the top-20 CadQuery API operation distribution across the         
benchmark, highlighting the dominance of \texttt{extrude} as the foundational
  operation while revealing a long-tail distribution of advanced operations
(e.g., \texttt{shell}, \texttt{fillet}, \texttt{chamfer},
\texttt{cutThruAll}) that progressively increases with difficulty level.
Representative samples from each level are showcased in
Fig.~\ref{fig:IterCAD-Draw}.

\noindent \textbf{IterCAD-Edit.}
The instruction-based editing track comprises $200$ test instances designed to evaluate localized parametric modifications. 
These instances are sourced from the Fusion 360 dataset, which inherently captures the complete, step-by-step intermediate construction history of solid parts. 
Leveraging this rich design sequence, we utilize a custom extraction script to translate the raw JSON-formatted modeling operations into executable CadQuery programs. 
To construct the benchmark, the natural-language editing instructions $\mathcal{T}_{\text{ref}}$ are generated by expert LLM based on the extracted operations. 
To guarantee absolute ground-truth correctness, the resulting target CadQuery programs $\mathcal{C}_{\text{tgt}}$ are rigorously executed and verified against their corresponding original geometries. 

Specifically, each verified evaluation instance provides: 
(i)~a source CadQuery program $\mathcal{C}_{\text{src}}$ representing an existing valid intermediate part, 
(ii)~a natural-language editing instruction $\mathcal{T}_{\text{ref}}$ specifying quantitative geometric modifications (e.g., adding prisms, cutting holes, adjusting dimensions), and 
(iii)~a target CadQuery program $\mathcal{C}_{\text{tgt}}$ with its corresponding ground-truth geometry for direct model comparison. 
The instructions are deliberately formulated with precise spatial coordinates and dimensional constraints to enable unambiguous geometric evaluation. 
Representative editing pairs are illustrated in Fig.~\ref{fig:IterCAD-Edit}, demonstrating the diversity of edit operations spanning feature addition, subtraction, and parametric adjustment.

\subsection{Metric Implementation Details}
\label{app:metric_details}

We provide the exact implementation details for the CD-TR curve and AUC-TR reported in the main text. For each evaluation record, the evaluator stores a top-level scalar \texttt{final\_cd}. This value is the Chamfer Distance computed by the CAD evaluator on the benchmark's normalized geometry representation; the curve-building script does not re-normalize or rescale it. Invalid generations remain part of the evaluation population: records with missing, non-finite, or negative \texttt{final\_cd} values (e.g., the failure sentinel \texttt{final\_cd = -1}) are kept in the denominator but never counted as recalled at any tolerance.

Formally, let $M$ denote the total number of evaluation records, including failed cases. Given a CD tolerance $\tau$, the CD-TR recall is
\begin{equation}
R(\tau)=\frac{1}{M}\sum_{i=1}^{M}\mathbb{I}\big(\mathrm{Valid}(y_i)\land CD_i\le \tau\big),
\end{equation}
where $CD_i$ is the stored \texttt{final\_cd} for case $i$. We sweep the tolerance on a logarithmic axis $x=-\log_{10}(\tau)$ over $x\in[1,5]$, where $x=1$ corresponds to the loosest tolerance $\tau=10^{-1}$ and $x=5$ to the strictest $\tau=10^{-5}$. In all reported results, the curve is evaluated using $401$ uniformly spaced points over $x\in[1,5]$, i.e., thresholds $\tau=10^{-x}$.

The scalar AUC-TR is computed by trapezoidal integration over this fixed log-tolerance interval and then normalized by the interval length:
\begin{equation}
\mathrm{AUC\text{-}TR}
=\frac{1}{x_{\max}-x_{\min}}\int_{x_{\min}}^{x_{\max}} R(10^{-x})\,dx.
\end{equation}
This normalization maps AUC-TR to $[0,1]$ in the main text and tables. A larger value therefore indicates that a method solves more cases and reaches lower CD tolerances over the entire evaluation set. By contrast, Mean CD and Median CD require no extra post-processing and are reported directly, as the underlying \texttt{final\_cd} is already computed based on models normalized to a unit-diagonal bounding box centered at the origin. For presentation clarity, all CD metrics in the main tables are scaled by $10^3$.

\begin{table*}[!t]
\centering
\setlength{\tabcolsep}{4pt}
\renewcommand{\arraystretch}{1.15}
\small
\resizebox{\textwidth}{!}{
\begin{tabular}{l l ccccc ccccc}
\toprule
 & & \multicolumn{5}{c}{\textit{\textbf{Direct Inference}}} & \multicolumn{5}{c}{\textit{\textbf{Agentic Workflow}}} \\
\cmidrule(lr){3-7} \cmidrule(lr){8-12}
\textbf{Models} & \textbf{Level} & IR\,(\%)\,$\downarrow$ & AUC-TR\,$\uparrow$ & Mean CD\,$\downarrow$ & Med. CD\,$\downarrow$ & Avg. Turn\,$\downarrow$ & IR\,(\%)\,$\downarrow$ & AUC-TR\,$\uparrow$ & Mean CD\,$\downarrow$ & Med. CD\,$\downarrow$ & Avg. Turn\,$\downarrow$ \\
\midrule
\multicolumn{12}{l}{\textit{Proprietary}} \\
\multirow{2}{*}{GPT-5}
 & Easy & 7.00  & 0.62 & 8.34  & 0.07 & 1.00 & 1.20  & 0.65 & 5.41  & 0.09 & 2.64 \\
 & Hard & 49.20 & 0.21 & 21.43 & 4.19 & 1.00 & 9.40  & 0.38 & 12.25 & 3.38 & 4.06 \\
\cmidrule(lr){1-12}
\multirow{2}{*}{Gemini-2.5-flash-lite}
 & Easy & 34.40 & 0.41 & 11.33 & 0.08 & 1.00 & 15.40 & 0.44 & 13.76 & 0.19 & 4.23 \\
 & Hard & 85.60 & 0.07 & 23.43 & 0.16 & 1.00 & 62.40 & 0.13 & 21.34 & 10.10 & 4.75 \\
\cmidrule(lr){1-12}
\multirow{2}{*}{Gemini-3-flash-preview}
 & Easy & 20.00 & 0.61 & 4.71  & 0.06 & 1.00 & 2.40  & 0.72 & 4.43  & 0.07 & 3.86 \\
 & Hard & 81.20 & 0.13 & 13.98 & 0.08 & 1.00 & 22.80 & 0.38 & 10.32 & 1.32 & 4.76 \\
\midrule
\multicolumn{12}{l}{\textit{Open-source}} \\
\multirow{2}{*}{GLM-4.6v}
 & Easy & 11.40 & 0.54 & 10.50 & 0.09 & 1.00 & 1.80  & 0.61 & 6.70  & 0.09 & 3.42 \\
 & Hard & 43.60 & 0.18 & 21.75 & 9.82 & 1.00 & 15.20 & 0.31 & 13.46 & 7.03 & 4.71 \\
\cmidrule(lr){1-12}
\multirow{2}{*}{InternVL3.5-30B-A3B}
 & Easy & 86.80 & 0.03 & 21.86 & 14.79 & 1.00 & 77.80 & 0.05 & 30.04 & 25.16 & 4.75 \\
 & Hard & 92.40 & 0.01 & 31.45 & 22.07 & 1.00 & 83.20 & 0.04 & 25.26 & 19.68 & 4.86 \\
\cmidrule(lr){1-12}
\multirow{2}{*}{InternVL3.5-8B}
 & Easy & 89.60 & 0.03 & 27.30 & 15.24 & 1.00 & 71.60 & 0.06 & 34.64 & 24.45 & 4.95 \\
 & Hard & 93.40 & 0.02 & 25.23 & 16.08 & 1.00 & 81.40 & 0.03 & 34.12 & 27.74 & 4.99 \\
\cmidrule(lr){1-12}
\multirow{2}{*}{Qwen3.5-35B-A3B}
 & Easy & 45.00 & 0.43 & 4.70  & 0.06 & 1.00 & 5.80  & 0.61 & 6.20  & 0.08 & 3.24 \\
 & Hard & 92.60 & 0.05 & 27.41 & 0.07 & 1.00 & 35.40 & 0.23 & 18.98 & 6.79 & 4.70 \\
\cmidrule(lr){1-12}
\multirow{2}{*}{Qwen3.5-4B}
 & Easy & 92.20 & 0.05 & 6.60  & 0.07 & 1.00 & 38.80 & 0.42 & 8.19  & 0.07 & 4.22 \\
 & Hard & 98.40 & 0.01 & 22.02 & 11.21 & 1.00 & 90.40 & 0.05 & 25.13 & 0.08 & 4.90 \\
\midrule
\multicolumn{12}{l}{\textit{Ours}} \\
\rowcolor{oursblue}
\cellcolor{white} & \cellcolor{oursblue}Easy & 1.80  & 0.69 & 6.75  & 0.06 & 1.00 & 0.20  & 0.74 & 2.49  & 0.06 & 2.77 \\
\rowcolor{oursblue}
\multirow{-2}{*}{\cellcolor{white}\textbf{\method{}}} & Hard & 11.20 & 0.39 & 13.06 & 2.66 & 1.00 & 0.40 & 0.47 & 7.70 & 1.78 & 3.17 \\
\bottomrule
\end{tabular}
}
\caption{Performance on the \textbf{\textit{IterCAD-Draw}} task across \textit{Easy} and \textit{Hard} subsets. Comparison between Direct Inference and Agentic Workflow.}
\label{tab:codepass_subset_difficulty}
\end{table*}

\begin{figure}[!t]
  \centering
  \includegraphics[width=0.98\linewidth]{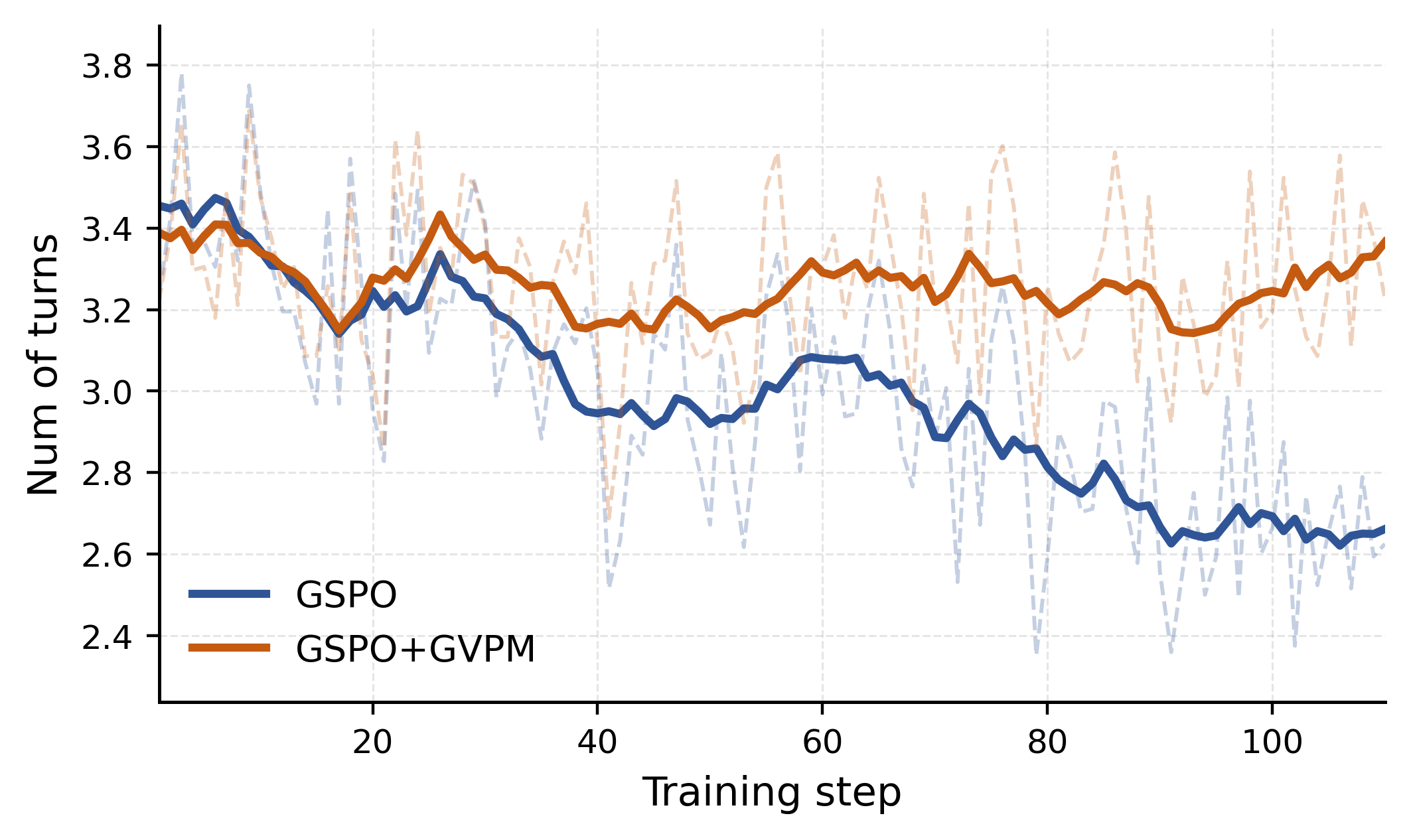}
  \caption{Average number of interaction turns during RL training. GSPO alone
   (blue) rapidly collapses toward single-turn behavior, indicating premature  
  termination. GSPO+GVPM (orange) maintains a higher and more stable turn
  count, reflecting substantive multi-turn refinement.}
  \label{fig:gspo_vs_gvpm}
  \vspace{-10pt}
\end{figure}                                                                                  

\begin{figure*}[!t]
  \centering                                                          
  \includegraphics[width=0.98\linewidth]{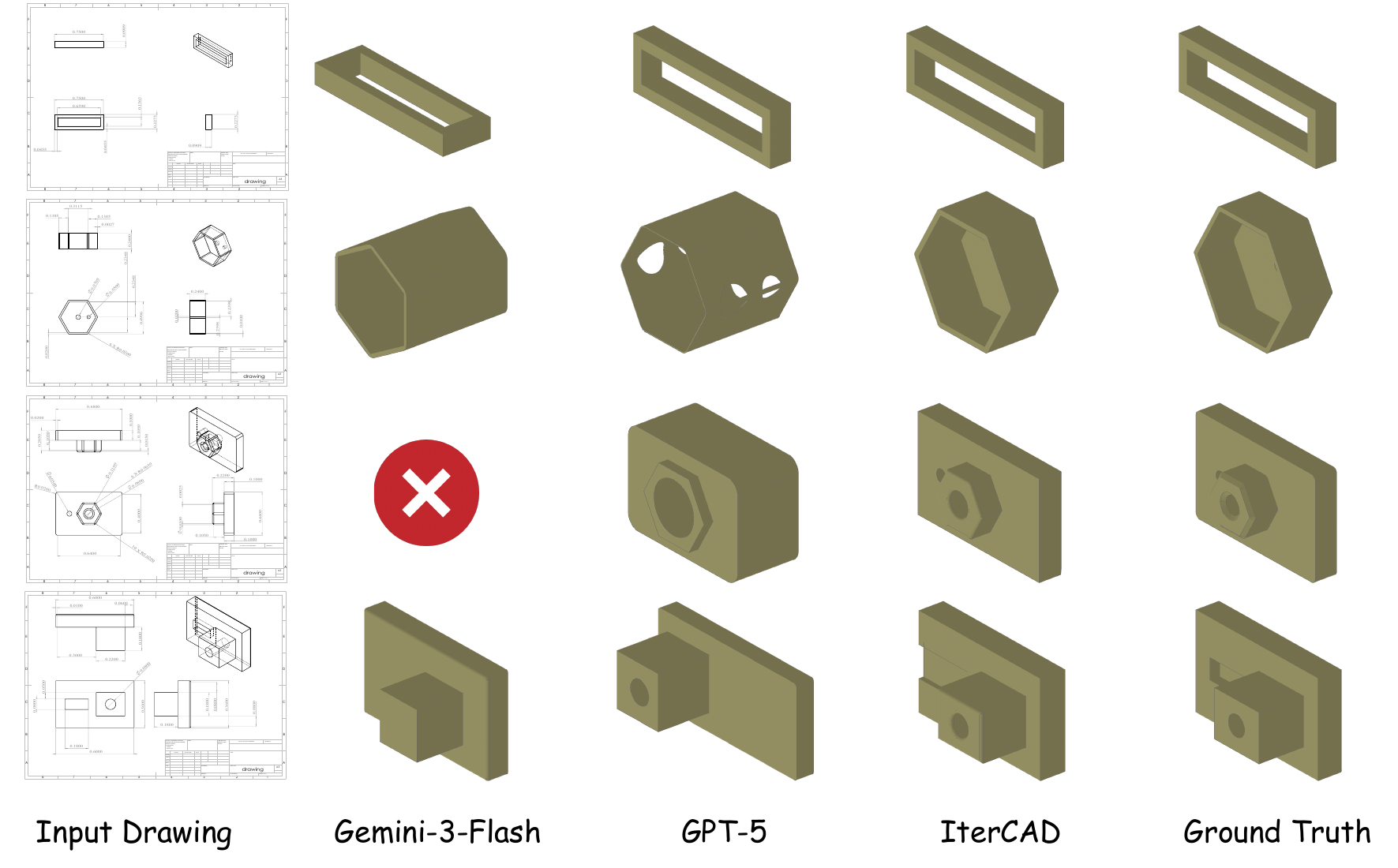}
  \caption{Qualitative comparison on representative \textit{IterCAD-Draw} examples. Given multi-view engineering drawings, \method{} produces geometries that more closely match the ground truth than strong proprietary baselines, especially for cases involving advanced features.}
  \label{fig:case-study}
  \vspace{-15pt}
\end{figure*}

\section{Additional Experimental Analysis}
\label{app:additional_exp_analysis}

We report all \textit{IterCAD-Draw} results under a pass@2 protocol.
As shown in Tab.~\ref{tab:codepass_subset_difficulty}, all evaluated models exhibit IR increases and AUC-TR decreases when transitioning from Easy to Hard instances, confirming that the difficulty stratification captures meaningful geometric complexity. Among baselines, even strong proprietary models suffer notable degradation on the Hard tier, where multi-operation reasoning and precise dimension extraction become critical. In contrast, \method{} maintains consistently low invalid ratios and high AUC-TR scores on both tiers. On the Hard subset,
where parts require intricate topological modifications such as shelling
combined with filleting and through-cuts, \method{} sustains a substantial
margin over models an order of magnitude larger, demonstrating that our
geometry-aware training recipe instills robust spatial reasoning rather than
merely memorizing simple extrusion patterns.

We further analyze the training dynamics of the RL stage by comparing the average number of interaction turns consumed during on-policy rollouts under GSPO alone versus GSPO augmented with Geometry-Viable Prefix Masking. As shown in Fig.~\ref{fig:gspo_vs_gvpm}, the two optimization variants exhibit markedly different behavioral trajectories. GSPO+GVPM maintains a consistently higher and more stable turn  
  count throughout training. By masking gradient contributions from
  geometry-stalled or error-cascading suffixes, GVPM ensures that the policy is
   neither rewarded for premature termination nor penalized for attempting
  productive refinement that ultimately fails in later turns. This
  credit-assignment correction encourages the agent to persist through
  multi-turn correction loops when the geometry warrants it, leading to the
  superior Mean CD ($5.09$ vs.\ $8.00$) and AUC-TR ($0.61$ vs.\ $0.58$)
  observed in Tab.~\ref{tab:ablation_components}.

\noindent \textbf{Visual feedback as observation.}
Rendered multi-view projections condition the policy as dialogue observations in both RL training and inference; they are not an extra reward term. Removing them from either stage degrades geometric fidelity (Tab.~\ref{tab:visual_feedback_ablation}), with the larger drop when they are withheld from RL rollouts.

\begin{table}[ht]
\centering
\setlength{\tabcolsep}{4pt}
\renewcommand{\arraystretch}{0.95}
\small
\begin{tabular}{lcc ccc}
\toprule
\textbf{Variant} & \textbf{Inf.} & \textbf{RL} & \textbf{IR}\,$\downarrow$ & \textbf{AUC-TR}\,$\uparrow$ & \textbf{Mean CD}\,$\downarrow$ \\
\midrule
No RL visuals & $\checkmark$ & -- & 0.90 & 0.57 & 8.42 \\
No inf. visuals & -- & $\checkmark$ & 0.40 & 0.60 & 6.62 \\
\rowcolor{oursblue}
\textbf{\method{}} & $\checkmark$ & $\checkmark$ & \textbf{0.30} & \textbf{0.61} & \textbf{5.09} \\
\bottomrule
\end{tabular}
\caption{Visual-observation ablation on \textit{IterCAD-Draw}. Inf. and RL indicate whether multi-view renderings are provided during inference and RL rollouts, respectively. IR is reported in percent.}
\label{tab:visual_feedback_ablation}
\end{table}

\noindent \textbf{Hard-subset failure analysis.}
We inspect all $500$ Hard \textit{IterCAD-Draw} cases, comparing ground-truth CadQuery programs with \method{}'s final multi-turn outputs. Using $CD>0.01$ as the failure criterion, the Hard failure rate is $17.2\%$ ($86/500$). Finishing and cutting operations are \emph{not} the main drivers: chamfered parts fail at $0\%$; shell and fillet fail at $4.2\%$ and $5.3\%$, both well below the Hard average. Failures instead concentrate in two modes: (i)~complex sketch geometry ($\sim$$30\%$), where dense \texttt{lineTo} contours or multiple \texttt{threePointArc}s are polyline-approximated, dropped, or topologically corrupted (failure rises from $11.7\%$ for $\texttt{lineTo}\le 4$ to over $30\%$ for complex contours); and (ii)~Boolean / multi-body collapse ($\sim$$37\%$), correlated with \texttt{union} (Hard failure $24.3\%$ with unions vs.\ $14.1\%$ without). In failed cut-containing cases the final code omits \texttt{cut} entirely in about two-thirds of instances, collapsing multi-step subtraction into a single extrude.

\section{Case Study}
\label{app:case_study}

\begin{figure*}[h]
  \centering                                                          
  \includegraphics[width=\linewidth]{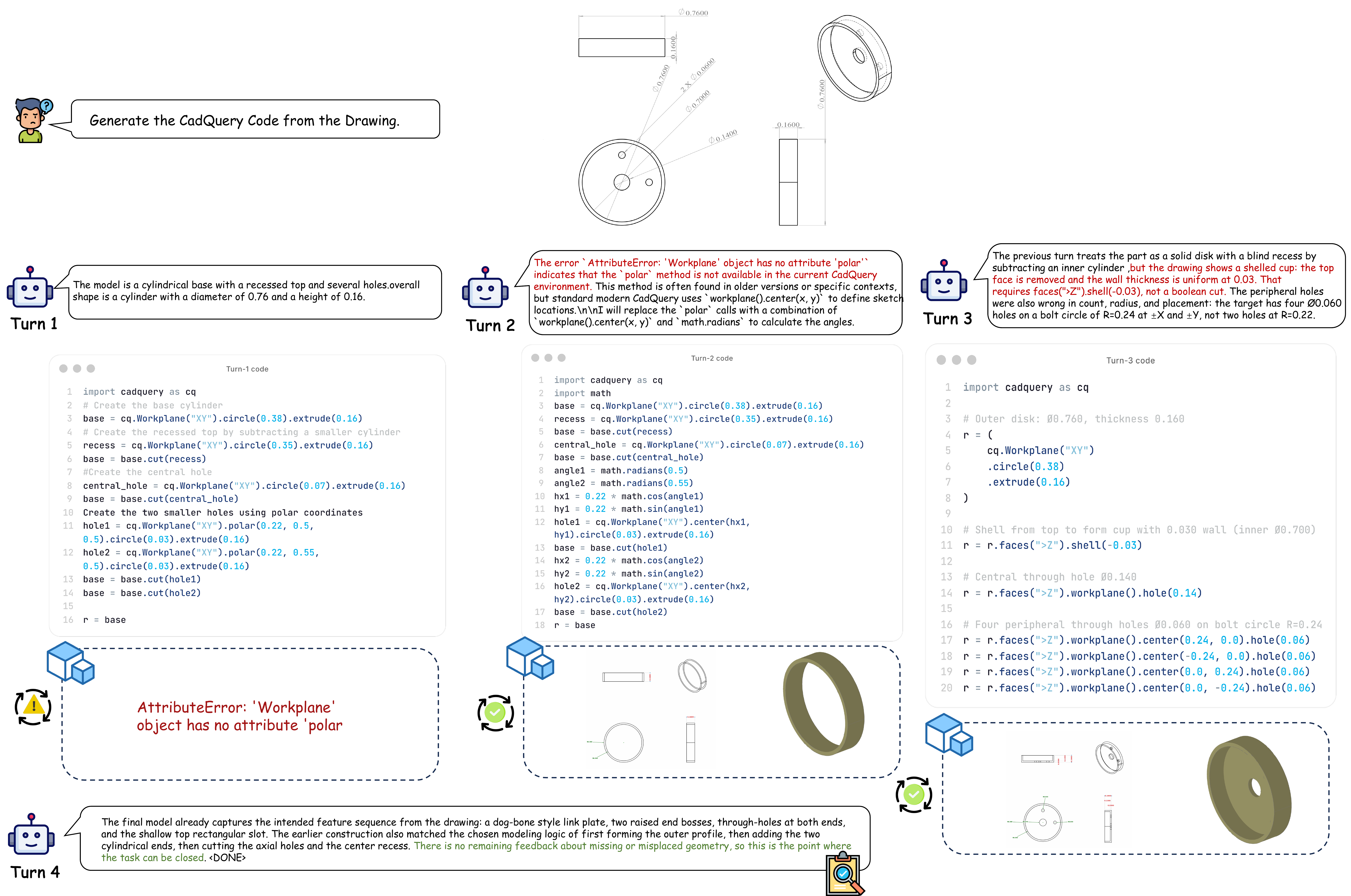}
  \caption{Drawing-to-code self-correction case. Starting from a dimensioned engineering drawing, \method{} first proposes an executable construction plan but encounters an unsupported CadQuery \texttt{polar} operation. Using sandbox feedback, the agent replaces the invalid call with explicit trigonometric placement and then refines the topology into a shallow circular plate with a center hole and four peripheral holes.}
  \label{fig:drawing-case}
  \vspace{-15pt}
\end{figure*}

\begin{figure*}[ht]
  \centering                                                          
  \includegraphics[width=0.8\linewidth]{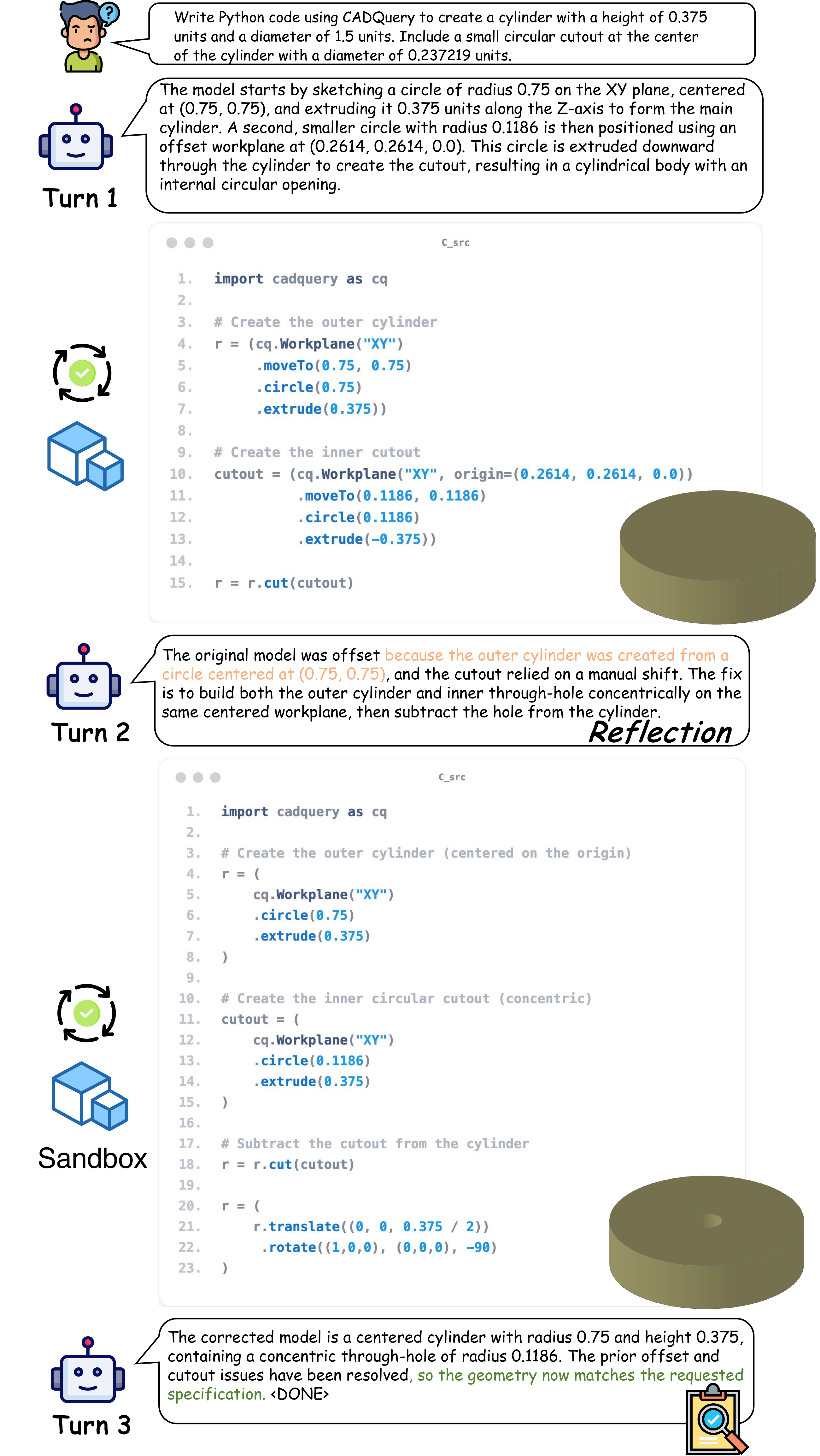}
  \caption{Text-to-CAD self-correction case. The initial code creates an offset cylinder and misses the intended concentric through-hole. After receiving sandbox feedback, \method{} revises the construction to center the outer cylinder and subtract a concentric circular cutout, producing the target geometry in the next turn.}
  \label{fig:text-case}
  \vspace{-15pt}
\end{figure*}

\begin{figure*}[ht]
  \centering                                                          
  \includegraphics[width=0.98\linewidth]{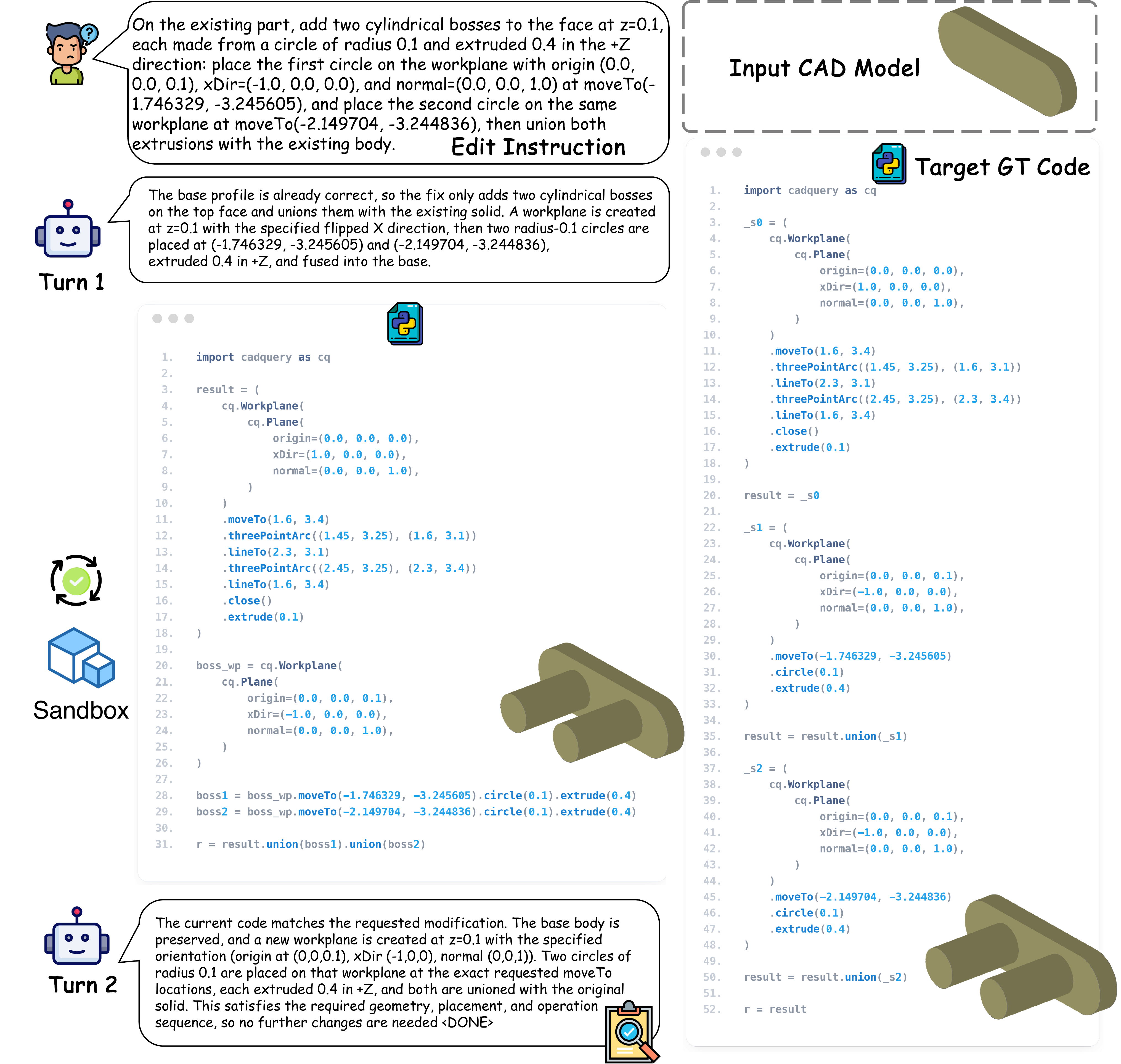}
  \caption{Instruction-based CAD editing case. Starting from an existing rounded base, \method{} follows the edit instruction to add two cylindrical bosses at specified positions and depths, while preserving the original base geometry and producing code consistent with the target program.}
  \label{fig:edit-case}
  \vspace{-15pt}
\end{figure*}

We provide qualitative case studies to illustrate how \method{} uses sandbox feedback to improve CAD code across different interaction modes. Fig.~\ref{fig:drawing-case} shows a drawing-to-code trajectory in which the first attempt fails because of an unsupported CadQuery API call, while later turns revise the construction strategy and recover the intended circular plate with a shallow shell, center hole, and peripheral holes. Fig.~\ref{fig:text-case} further demonstrates text-conditioned self-correction: after detecting an offset cylinder and a missing concentric through-hole, the agent rewrites the program around a centered workplane and obtains the requested geometry. Fig.~\ref{fig:edit-case} highlights localized editing, where \method{} adds two cylindrical bosses while preserving the original base geometry.

\begin{figure*}[ht]
  \centering                                                          
  \includegraphics[width=0.98\linewidth]{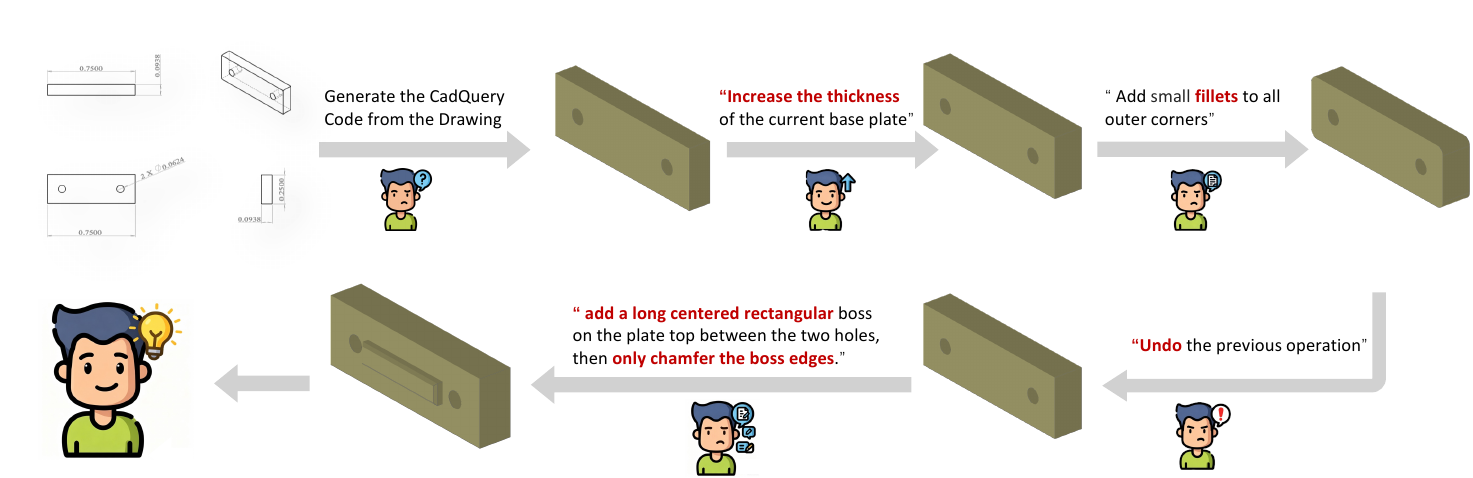}
  \caption{Unified generation-and-editing example. \method{} first reconstructs a base plate from the input drawing, then sequentially applies user instructions to increase thickness, add fillets, undo an operation, and insert a centered rectangular boss with chamfered edges, demonstrating flexible multi-turn design control across generation and editing.}
  \label{fig:CAD-gen_edit}
  \vspace{-15pt}
\end{figure*}

Beyond these single-task examples, Fig.~\ref{fig:CAD-gen_edit} illustrates the unified generation-and-editing ability of \method{} within one continuous design session. The agent first reconstructs a base CAD model from an engineering drawing, then follows subsequent natural-language instructions to modify thickness, add fillets, undo an operation, and introduce a new chamfered boss. This example shows that \method{} is not limited to isolated generation or one-step repair; instead, it can maintain design state across turns and support controllable CAD evolution through mixed generation and editing commands.

\section{Prompt}
The system prompt we used for training and evaluation is shown in Figure \ref{fig:cad_system_prompt}.

\begin{figure*}[t]
\centering
\begin{tcolorbox}[
     title=\textbf{System Prompt for IterCAD Reasoning},
     colback=gray!5,
     colframe=gray!60!black,
     boxrule=0.8pt,
     fonttitle=\bfseries,
     left=1em, right=1em, top=1em, bottom=1em,
     width=0.98\textwidth
]
\small

\textbf{Role.}
You are an expert CAD engineer and Python programmer specialized in CadQuery.
Given visual, textual, code, and optional feedback inputs, your task is to generate or modify a 3D CAD model one reasoning step at a time.

\vspace{0.5em}
\textbf{Input Sources.}
The input may include one or more of the following:
\begin{enumerate}[leftmargin=*, noitemsep, topsep=2pt]
    \item \textbf{Technical Drawing Image}: orthographic projections such as Front, Top, Side, and ISO views with dimensions.
    \item \textbf{Text}: modeling instructions, dimensional constraints, or edit requests.
    \item \textbf{Existing Code}: a CadQuery script that should be preserved or modified when possible.
\end{enumerate}

\vspace{0.5em}
\textbf{Objective.}
Create or edit a 3D model that satisfies the user request.

\vspace{0.5em}
\textbf{Output Format.}
Always start with \texttt{\textless thinking\textgreater} and output one of the following:

\begin{enumerate}[leftmargin=*, itemsep=4pt, topsep=2pt]
    \item Structure the \texttt{\textless thinking\textgreater} process strictly based on the provided inputs:
    \begin{itemize}[leftmargin=1.5em, noitemsep, topsep=2pt]
        \item \textbf{If no feedback is provided}, such as initial generation or completely new instructions:
        \begin{itemize}[leftmargin=1.5em, noitemsep]
            \item \textbf{Requirement Analysis}: break down visual or textual inputs into CadQuery features.
            \item \textbf{Plan}: define the origin, workplanes, sketch sequence, Boolean operations, and key dimensions.
        \end{itemize}

        \item \textbf{If feedback or errors are provided}, such as fixing a previous attempt:
        \begin{itemize}[leftmargin=1.5em, noitemsep]
            \item \textbf{Feedback Analysis}: precisely identify what failed, including compilation errors, missing geometry, and other issues.
            \item \textbf{Modification Plan}: state the targeted local edits required and explicitly define what existing code must be preserved.
        \end{itemize}

        \item \textbf{For all scenarios}, always conclude the thinking process with:
        \begin{itemize}[leftmargin=1.5em, noitemsep]
            \item \textbf{Precision Check}: verify numerical values and confirm all 2D profiles are closed before extrusion.
        \end{itemize}
    \end{itemize}

    Then output the CadQuery implementation in the following format:

    \vspace{0.3em}
    \noindent\hspace*{1em}%
    \begin{minipage}{0.92\linewidth}
    \ttfamily\footnotesize
    \textless/thinking\textgreater\\
    \textasciigrave\textasciigrave\textasciigrave python\\
    import cadquery as cq\\
    \# ... your code ...\\
    r = ...\\
    \textasciigrave\textasciigrave\textasciigrave
    \end{minipage}

    \item If feedback explicitly confirms that the 3D model is correct with no remaining issues, briefly state the assessment in \texttt{\textless thinking\textgreater \textless/thinking\textgreater}, then output \texttt{\textless DONE\textgreater}.
\end{enumerate}

\textbf{Code Implementation Rules.}
\begin{itemize}[leftmargin=*, noitemsep, topsep=2pt]
    \item Use Python as the programming language.
    \item Use CadQuery with \texttt{import cadquery as cq}.
    \item Assign the final result to variable \texttt{r}.
    \item If scaling is needed, define \texttt{scale\_factor} and multiply dimensions explicitly. Never use \texttt{.scale()}.
    \item Ensure all 2D profiles are valid and closed before extrusion when required.
    \item Do not use visualization calls such as \texttt{show\_object()}.
    \item Preserve correct existing geometry when editing code.
    \item Do not rewrite unrelated parts of the code if only local edits are needed.
\end{itemize}

\end{tcolorbox}
\caption{System prompt for IterCAD code generation and editing.}
\label{fig:cad_system_prompt}
\end{figure*}

\end{document}